\documentclass{CVM}
\usepackage{tabularx}
\usepackage{makecell}
\usepackage{color}

\CVMsetup{
	type      = {Research Article},
	doi       = {s41095-0xx-xxxx-x},
	title     = {
	
		6DOF Pose Estimation of a 3D Rigid Object based on Edge-enhanced Point Pair Features},
	author    = {Chenyi Liu$^{1}$, Fei Chen$^{2}$\cor{}, Lu Deng$^{3}$, Renjiao Yi$^{1}$, Lintao Zheng$^{4}$, Chenyang Zhu$^{1}$, Jia Wang$^{5}$, Kai Xu$^{1}$\\
	},
	runauthor = {Chenyi Liu et al.},
	abstract  = {
		
		  The point pair feature (PPF) is widely used for 6D pose estimation. In this paper, we propose an efficient 6D pose estimation method based on the PPF framework.  We introduce a well-targeted down-sampling strategy that focuses more on edge area for efficient feature extraction of complex geometry. A pose hypothesis validation approach is proposed to resolve the symmetric ambiguity by calculating edge matching degree. We perform evaluations on two challenging datasets and one real-world collected dataset, demonstrating the superiority of our method on pose estimation of geometrically complex, occluded, symmetrical objects. We further validate our method by applying it to simulated punctures.

	},
	keywords  = {point pair feature; pose estimation; object recognition; 3D point cloud},
	copyright = {The Author(s) 2022},
}

\begin{document}
	
	\maketitle
	
	\begin{figure}[b!] \vskip -4mm
		\small\renewcommand\arraystretch{1.3}
		\begin{tabular}{p{81mm}} \toprule\\ \end{tabular}
		\vskip -7.0mm \noindent \setlength{\tabcolsep}{1pt}
		\begin{tabular}{p{3.5mm}p{80mm}}
			&\\[-3mm]    
			$1\quad $ &College of Computer, National University of Defense Technology, Changsha 410073, China. E-mail: C.Liu, liuchenyi\_1013@nudt.edu.cn; R.Yi, yirenjiao@nudt.edu.cn; C.Zhu,  zhuchenyang07@nudt.edu.cn. K.Xu,
			kevin.kai.xu@gmail.com. \\
			$2\quad $ & Department of Spine Surgery, The Second Xiangya Hospital, Central South University, Changsha 410011, China. E-mail: F.Chen, chenfei1972@csu.edu.cn\cor{}. \\
			$3\quad $ & Clinical Nursing Teaching and Research Section, The Second Xiangya Hospital, Changsha 410011, China. E-mail: L.Deng, csdenglu1026@csu.edu.cn.\\
			$4\quad $ &College of Meteorology and Oceanography, National University of Defense Technology, Changsha 410073, China. E-mail: L.Zheng,  zhenglintao13@nudt.edu.cn.\\
			$5\quad $ &Beijing Institute of Tracking and Communication Technolog,  Beijing 100094, China. E-mail: J.Wang,  kelexuebi2009@163.com.\\
			&\hspace{-5mm} Manuscript received: 2022-05-11; \vspace{-2mm}
		\end{tabular} \vspace {-3mm}
	\end{figure}
	
	\begin{figure}[t!]
		\centering
		\includegraphics[width=0.5\textwidth]{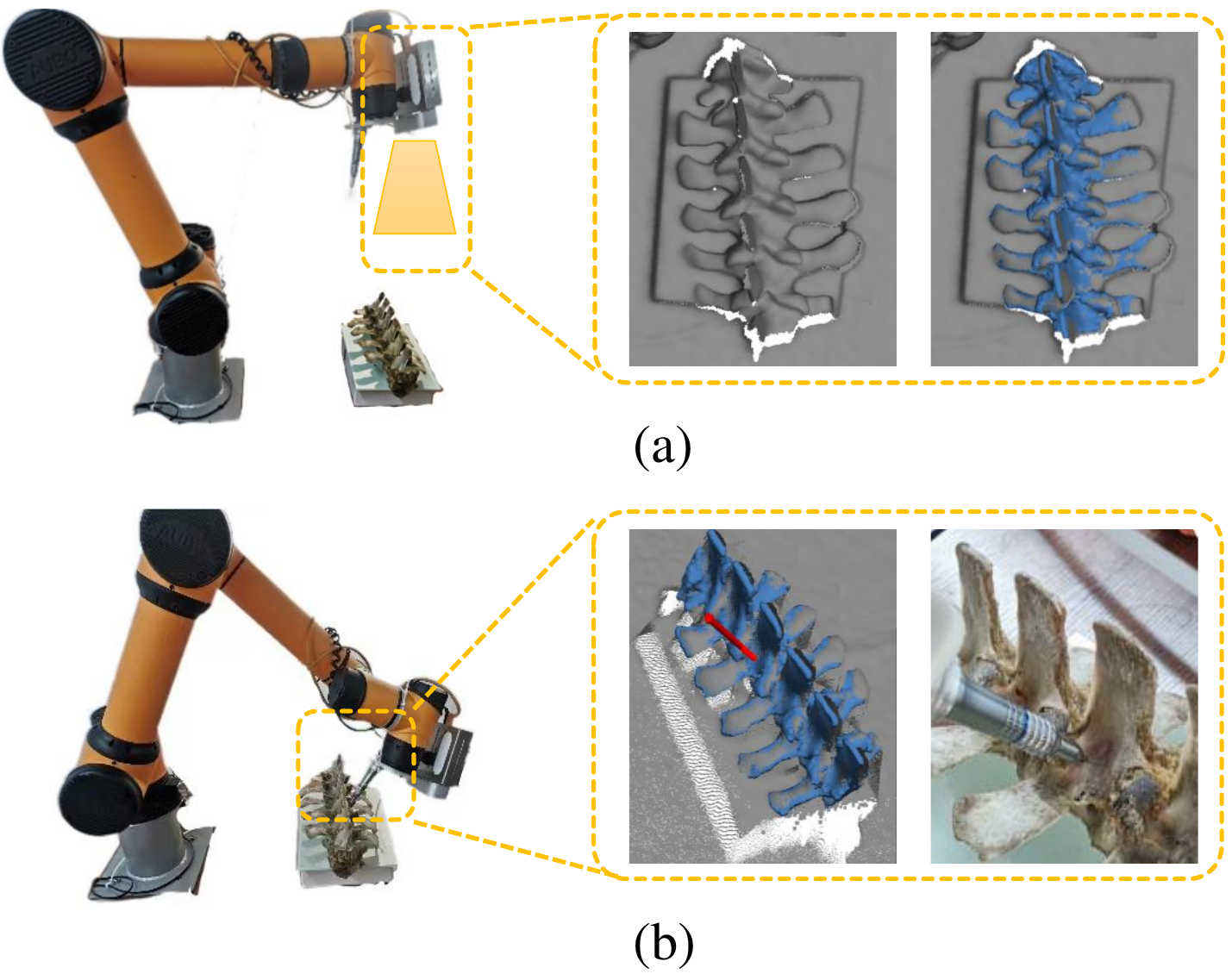}
		\caption{\label{fig:overview} Our experiment for robot-operated positioning with vision-based navigation. (a) The depth camera scans the spine for template-based pose estimation. (b) After matching, the robotic arm points and drills the spine with a predetermined pose and position.}
	\end{figure}
	\section{Introduction}

The goal of 6D pose estimation is to detect the position and orientation of a target object to obtain a rigid transformation from the object coordinate system to the camera coordinate system. Pose estimation has been considered an important part of target recognition and scene understanding. Pose estimation has also been widely used in industrial and medical fields. In the medical field, with the continuous development of medical imaging, computer-assisted surgery technology, and 3D vision technology, 3D vision-based navigating robot-operated surgery has become a trend \cite{li2019mixed,wang2021current}. In 3D vision-based navigating robot-operated surgery, the registration of preoperative 3D models reconstructed by medical imaging and intraoperative spine point clouds acquired by depth cameras is crucial.

	In real surgical scenarios, the human spine has a complex geometry and features of high occlusion and symmetry  \cite{kim2017vertebrae}, thus potentially leading to algorithmic miscalculations. There is no satisfactory and universal solution for this problem. In this work, we propose a method of pose estimation for special geometries of the spine. For the complex shape of the spine, we found that more spine feature points exist on the edges. Therefore, an edge-focused sampling method is used to select stable and salient points to generate stable transformation hypotheses. For the ambiguity of spinal symmetry, we consider that the difference in details between symmetric and highly occluded objects can be effectively distinguished by the degree of edge matching.
	
	Overall, the contributions of our work are summarized as follows.
	\begin{itemize}
		\item A well-target down-sampling strategy combines edge information. It effectively retains edge points and points with large curvature variations. Robust hypothesis generation is achieved by sampling stable feature points.
		
		\item A pose hypothesis verification method considers the degree of matching with edge points. It has an early exit strategy to reduce time costs.

		\item An
		experimental platform of robot-operated positioning based on this method is implemented. We use the position-based visual servoing scheme to control the robot arm to reduce the deviation of the drilling position.
	\end{itemize}
	
	\section{Related works}
	This section reviews the correlation algorithms of pose estimation in 3D point clouds, point pair features and their modifications. 
	
	\subsection{Pose estimation methods}

	Many different methods have been proposed for 3D object detection and pose estimation. Existing research methods can be roughly divided into feature-based methods, template matching methods, point-based methods, and Deep learning-based methods. Feature-based methods can be considered the broadest solution, roughly divided into global feature-based and local feature-based algorithms.  The algorithm based on global features  \cite{5651280, marton2011combined, 6385874} has good performance in calculating time and memory consumption. However, the algorithm is limited in clinical applications due to its sensitivity to occlusion and noise, and the need to pre-isolate the region of interest from the background. The algorithm based on local features \cite{ johnson1999using, 5152473, tombari2010unique, rusu2009detecting} is more robust to occlusion and clutter. Nevertheless, it will lead to additional computation time during the subsequent matching and hypothesis validation, so it does not meet the requirements of a real-time surgical navigation system. The method based on template matching  \cite{hinterstoisser2011multimodal} can detect texture-free targets but is sensitive to surgical instrument occlusion. The main application of the point-based method is the Iterative Closest Point algorithm (ICP)  \cite{besl1992method} and its variants  \cite{chen1992object, rusinkiewicz2001efficient}. The ICP algorithm and its variants are dependent on the initial pose and are usually used in pose refinement. Deep learning-based methods \cite{park2019pix2pose, hodan2020epos,liang2019pointnetgpd,lan2022arm3d,zeng2021ppr} perform well in public 3D datasets. However, deep learning-based methods require significant computational power and time to label datasets. The difficulty of collecting medical samples and the small amount of data hinders the application of deep learning-based methods for surgical navigation.
	
	\subsection{Point pair feature}
	In 2010, Drost et al. \cite{drost2010model} proposed a rigid 6D pose estimation method based on point pair feature (PPF), which is a compromise between local feature and global feature methods, striking a good balance between accuracy and speed. PPF describes the surface of an object through global modeling of four-dimensional features defined by directional point pairs. These features are used to find the corresponding relationships between scene and model point pairs, generate numerous candidate hypotheses, and then cluster and sequence the candidate poses to obtain the final hypotheses. PPF features are low-dimensional features of the oriented points and are suitable for objects with a rich surface variation. 
	Moreover, the PPF descriptors with global significance have stronger discriminative power than most local features. It is suitable for the complex structure and occluded objects studied in this paper, so we choose the PPF framework as the backbone.
	
	\par Because of the advantages of PPF, many improvement schemes based on PPF have been proposed. Choi et al. \cite{choi2016rgb} proposed a color point pair feature (CPPF), which uses color information to significantly improve the discrimination and accuracy of traditional point pair features. Drost et al.  \cite{drost20123d} proposed the concept of geometric and textured edges. Geometric edges are obtained using the intensity image and depth image to construct multimodal point pair features. Liu et al. \cite{liu2018point} proposed a novel descriptor named Boundary-to-Boundary-using-Tangent-Line (B2B-TL) to estimate the pose of industrial parts. Vock et al. \cite{ vock2019fast} utilized point pair features that are on edge for the quick generation of transformation guesses in a Random Sample Consensus setting. Inspired by the above article,  we propose a down-sampling method using a combination of edge points and geometric high curvature feature points for the spine. A pose hypothesis verification method based on edge matching is proposed to make it more competitive in detecting geometrically complex and symmetrical objects such as the spine.

	\par The rest of this paper is organized as follows. Section \ref{sec:ppfmethod} describes the original PPF method, and Section \ref{sec:PROPOSEDALGORITHM} describes our proposed method and the design 
	of  
	robot-operated positioning experiments. 
	Experimental results for the spine dataset and the public datasets are given in Section \ref{sec:Experiments}. Section \ref{sec:Conclusion} concludes the paper.
	
	\section{PPF Method}
	\label{sec:ppfmethod}
	Our approach is based on the original PPF method \cite{drost2010model}. To better understand this article, we will introduce the basic framework of this method in this section.
	\begin{figure}[t!]
		\centering
		\includegraphics[width=0.4\textwidth]{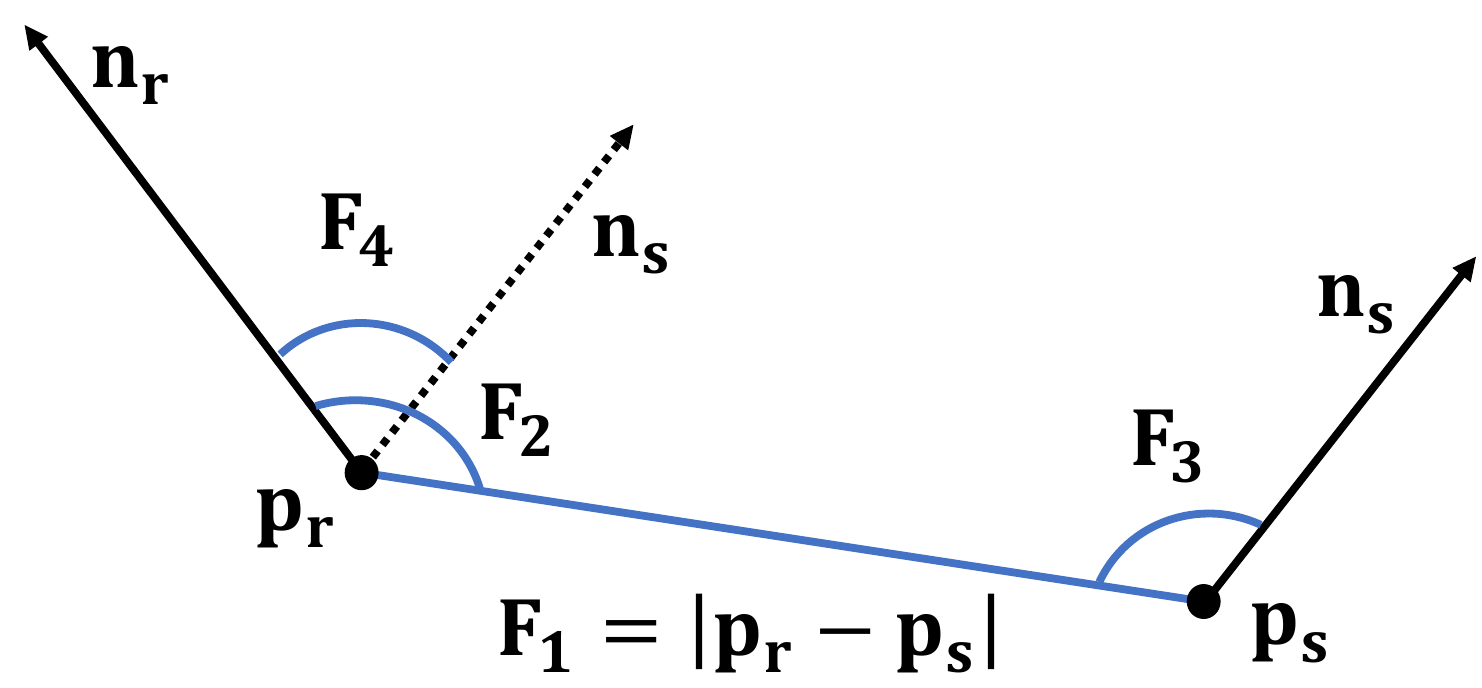}
		\caption{\label{fig:1} The point pair feature definition.}
	\end{figure}
	\begin{figure*}[t!]
		\centering
		\includegraphics[width=2.05\columnwidth]{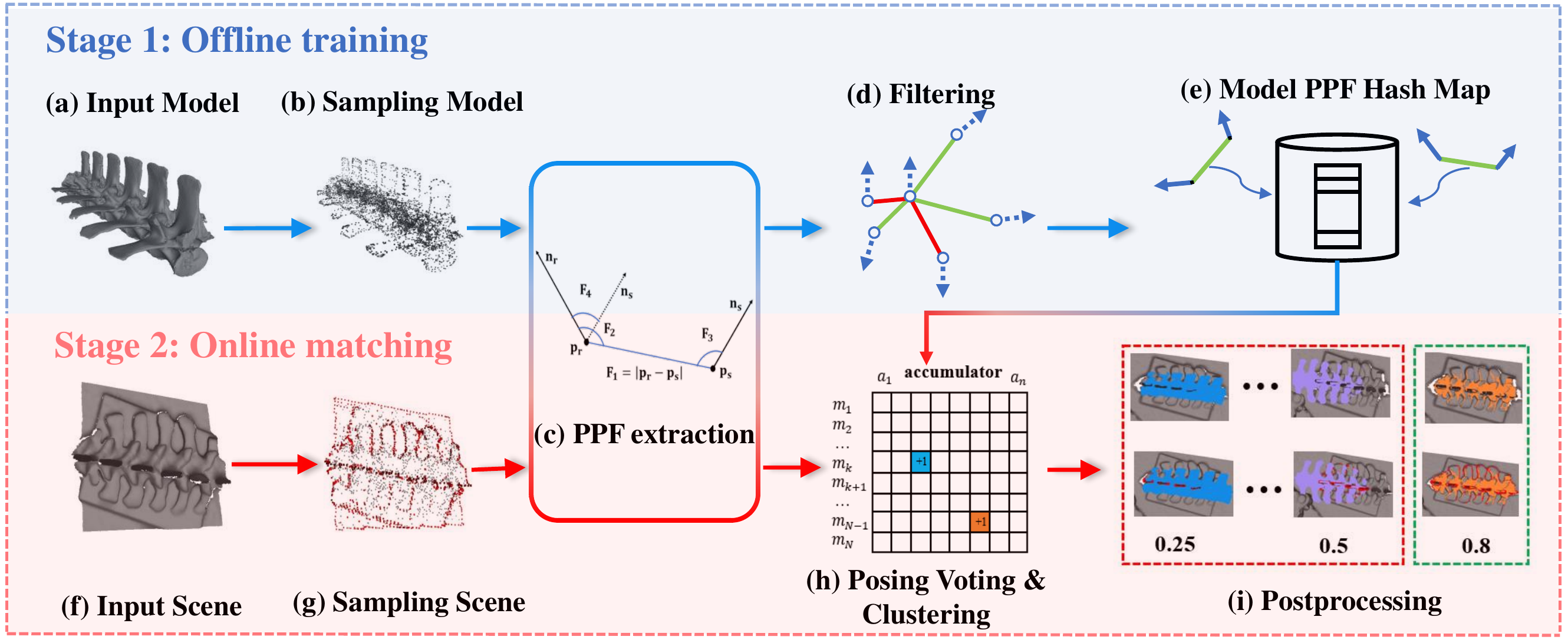}
		\caption{\label{fig:2}  The framework of the proposed method. It is mainly divided into two stages: offline training and online matching. In the offline training stage, the CAD model is input  (a)  . After downsampling (b), the PPF features are extracted from the model (c). In the processing of the filtering (d), we filter out PPF features with angles higher than $ 175^{\circ} $ or lower than $ 5^{\circ} $ by judging the normal vector angle between the point pairs. PPF features are extracted and stored in a hash table (e). In the online matching stage, the scene point cloud is input (f), In the pre-processing of the scene point cloud (g), we use a clustered down-sampling method that takes into account the normal vector information, and focus on the edge point cloud and the points with large curvature. The PPF features extracted from the scene point cloud (c) are matched to the hash table, and the candidate poses are generated by voting and pose clustering(h). Each candidate pose is then post-processed(i). The pose with the highest matching score is selected by an improved edge-based pose verification method. Finally, we use ICP to refine the final result pose.
	}
	\end{figure*}
	\subsection{Point pair feature}
	The point pair feature is used to describe the relative distance and normals of a pair of oriented points, as shown in Fig. \ref{fig:1}.  Given a reference point $ p_r $ and a second point $ p_s $ with normal  $ n_r $ and  $ n_s $ respectively, the PPF is a four-dimensional vector which is defined as:
	\begin{equation}\label{eq:PPF}
		\operatorname{PPF}\left(\mathbf{p}_{r}, \mathbf{p}_{s}\right)=\left(\|\mathbf{d}\|_{2}, \angle\left(\mathbf{n}_{r}, \mathbf{d}\right), \angle\left(\mathbf{n}_{s}, \mathbf{d}\right), \angle\left(\mathbf{n}_{r}, \mathbf{n}_{s}\right)\right),
	\end{equation}
	where  $ \mathbf{d} =\mathbf{p}_r - \mathbf{p}_s $, $\angle (\mathbf{a}, \mathbf{b}) $ is the angle between the vector $ \mathbf{a} $ and the vector $ \mathbf{b} $.
	
	\subsection{Drost's pipeline}
	
	The PPF method can be divided into offline global modeling and online local matching. 
	
	\par 
	In the offline global modeling phase, to create a description of the model, the model is down-sampled using uniform sampling. Then the point pair features are computed and quantified for all permutations of model point pairs. The point pair features are made to be stored as hash keys in a hash table by the quantization function, and the value encodes the pose of the feature relative to the model. The pose of the model is encoded by storing the index of the reference point $ p_r $ and an angle $ a_m $, the latter of which represents the angle between the projection of the model point pair concatenation and the positive direction of the Y-axis.

	\par
	The online local matching phase consists of two parts: (1) find the correspondence between point pairs using four-dimensional point pair features; (2) generate hypothetical poses from the correspondences and then cluster to obtain the best object pose. In the first part, the reference points are sampled from the scene. Uniform down-sampling of the scene point cloud is performed to obtain a set of scene points, and then the $i$-th (default  $ i = 5 $) scene point is used as the reference point. Make this reference point calculate the PPF together with all other scene points. And map it to the model reference point and angle $\alpha_m $ by matching using the previously constructed hash table. This process effectively solves the correspondence problem between point pairs by matching point pairs with the same quantized PPF. In the second part, the  $ \alpha_s $ of the scene point pairs are calculated.  $ \alpha_s $ represents the angle between the connected projection of the scene point pairs and the positive direction of the Y-axis. For each matched point pair feature, the angle $ \alpha = \alpha_m-\alpha_s $, and then voting is performed in the Hough space of  $ (p_r, \alpha) $. The maximum value of the number of votes in the Hough space is extracted to form a pose hypothesis. After the valid candidates are generated for all reference points, cluster the similar poses grouped by judging the rotation and translation that do not exceed the threshold. The group with the highest cumulative number of votes is the resulting pose hypothesis.

	\section{PROPOSED ALGORITHM}
	\label{sec:PROPOSEDALGORITHM}
	\subsection{ Overview of Our Approach}
	
	We propose a new 6D pose estimation algorithm, the specific framework of which is shown in Fig. \ref{fig:2}. Based on PPF, we mainly make the following improvements. First, for the pre-processing operation of the input model, we filter out the point pair features that tend to interfere with the matching based on the normal vector angle for the input model. Secondly, for the pre-processing operation of the scene point cloud, we use a clustered down-sampling method that preserves the edge point cloud. Finally, the pose verification operation is performed by checking the matching degree of edges to filter out wrong poses. The proposed improvements are described in the following sections.

	\begin{figure}[b!]
		\centering
		\includegraphics[width=0.35\textwidth]{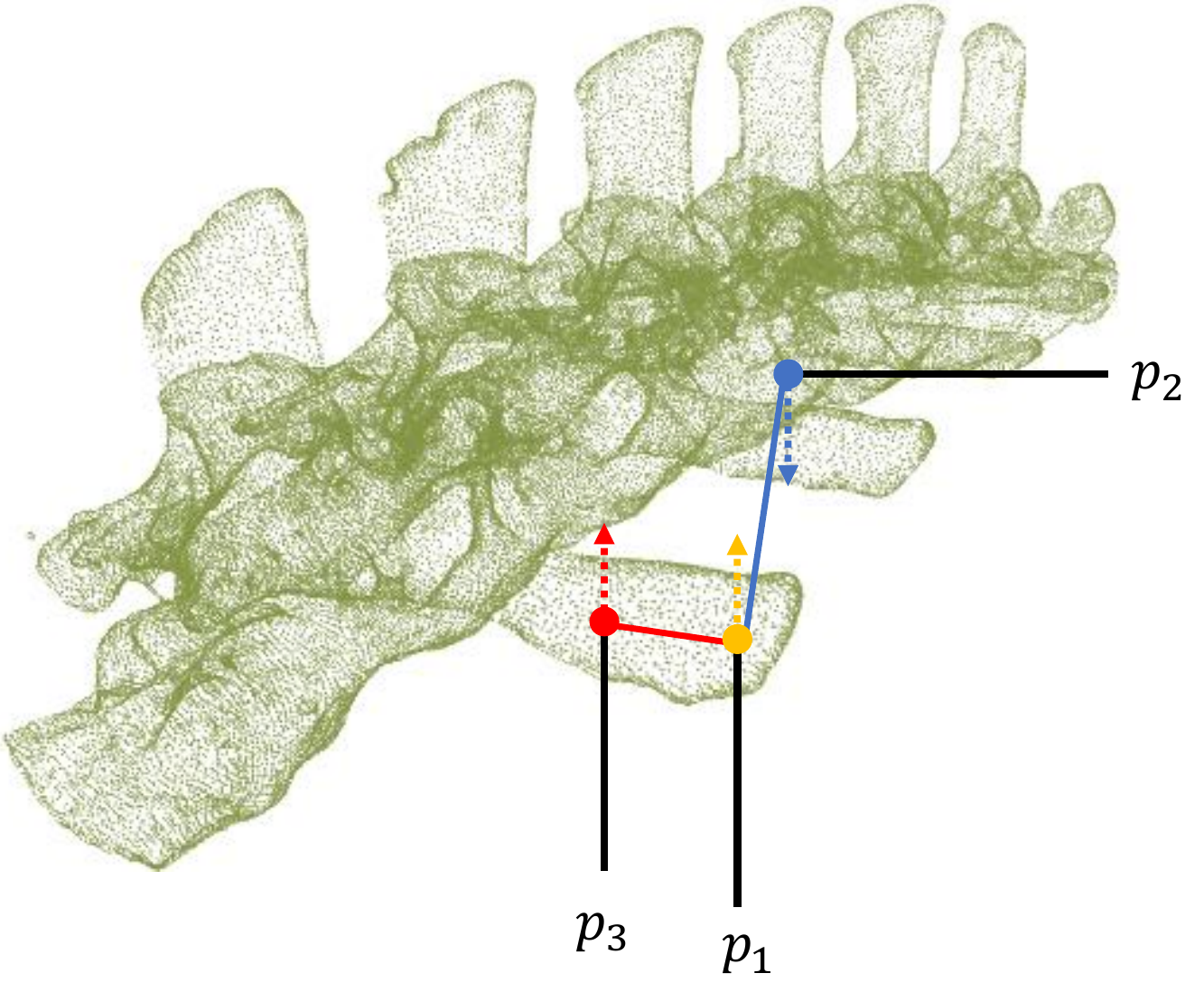}
		\caption{\label{fig:3} When $ p_1 $ is used as the reference point, $ p_2 $ which has a normal vector angle of more than $ 175^{\circ} $ with $ p_1 $, will not appear in the same view due to the visibility constraint of the viewpoint. Due to the specificity of the plane structure, the points in the same plane such as $ p_3 $ are easily mapped to the same hash bin in the hash table, which reduces the performance of the algorithm.}
	\end{figure} 
	
	\subsection{Offline training}
	In the offline training phase, all point pair features of the model are extracted and stored in a hash table to create a global model description. However, due to self-blocking, the global description contains some redundant point pair features that never appear in the input scene. The redundant point pair features not only increase the search time in the online matching phase but also increase the matching error. To mitigate the negative impact of redundant point pair features, we adopt a method based on \cite{lu39three} to determine the visibility problem of point pair features by using the normal vector angle between point pairs. If the angle between the normal vectors of two oriented points is higher than $175^{\circ}$, we consider the point pair as almost invisible. Therefore, the storage of point pair features is not performed.

	On the other hand, it is common for the traditional PPF method to degrade when the object has many repetitive features, such as large planes. Therefore, we do not store the normal vector angle of two oriented points less than $5^{\circ}$, so the algorithm focuses more on the geometrically-rich point pair features. As shown in Fig. \ref{fig:3}, we mainly filter out the points that are self-obscured by the viewpoint and the points that are on the same plane.

	\subsection{Online matching}
	\begin{figure}[b!]
		\centering
		\includegraphics[width=1\columnwidth]{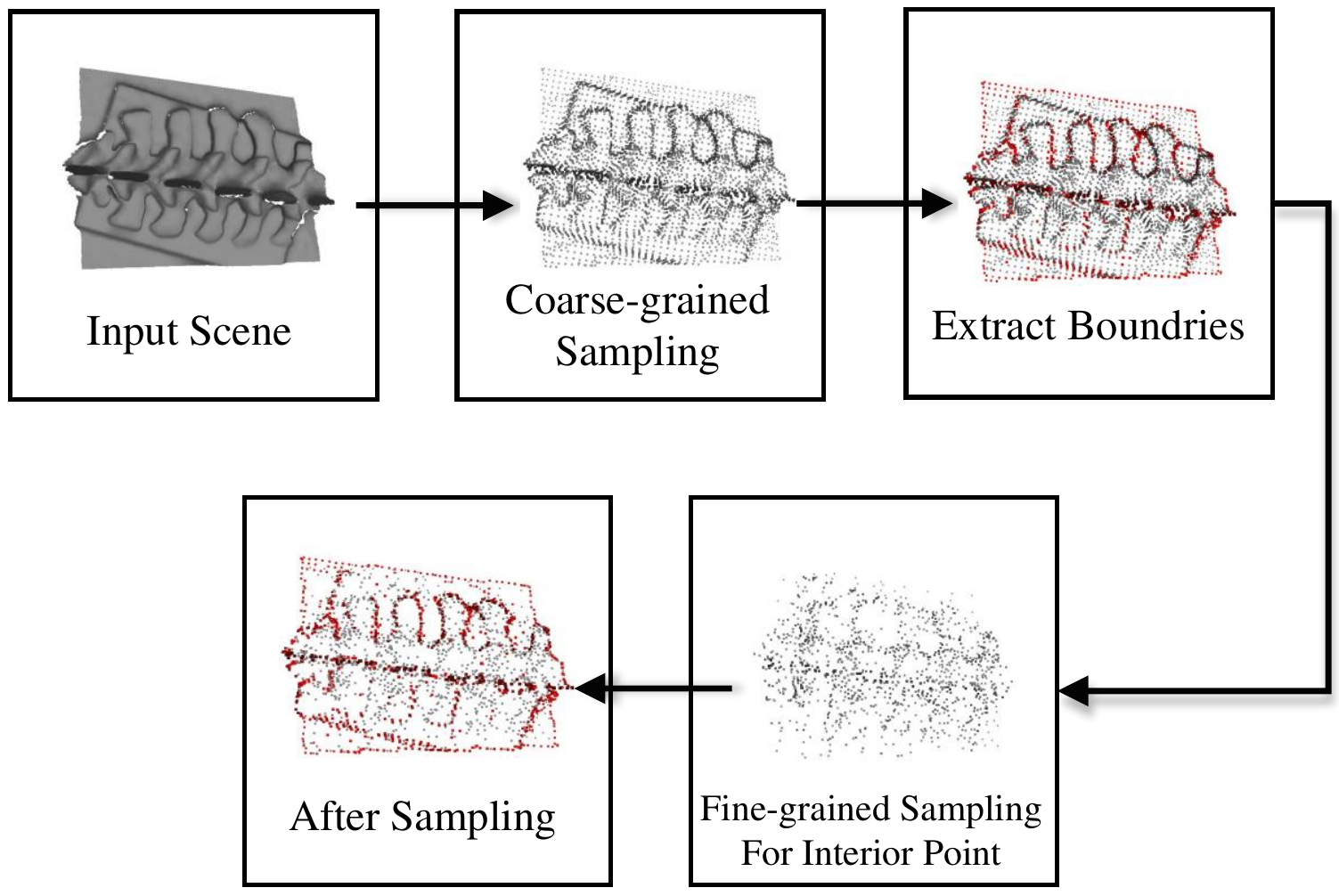}
		\caption{\label{fig:4} The flow chart for clustered down-sampling method considering edge information.}
	\end{figure}
	\subsubsection{Pre-processing}
	In order to accelerate the computation of object poses, the scene point cloud must be down-sampled. Unlike Drost's method \cite{drost2010model}, we use a clustering down-sampling method that takes into account normal, similar to  \cite {vidal2018method, guo2021efficient}. However, We also focused on the edge points of the point cloud. Edge points can robustly describe the shape of the object, and for complex objects such as spinal bones, feature points have a higher probability of being presented at the edges. Our approach is shown in Fig. \ref{fig:4}, where we first create a multi-resolution grid structure to discretize the scene point cloud according to the diameter of the model. Similar points with normal angles difference less than  the threshold $ \theta $ are then merged in a voxel grid. After the first fine-grained sampling, we extract the edge point clouds and continue with a fine-to-coarse multi-resolution  sampling strategy for the non-edge points. To prevent some geometric features from being filtered out in the coarse-grained grid, the threshold $\theta $ is gradually reduced proportionally. The above operation can effectively preserve the edge points and the points with large curvature.

	\subsubsection{Feature extraction}

	For scene point clouds, we follow the solution proposed in  \cite{drost2010model}, choosing 1/5th  of points in the scene as reference points and other points as the second point in the point pair feature. To improve the efficiency of the matching part, we use the KD-tree structure and adopt the intelligent sampling strategy of Hinterstoisser et al. \cite{hinterstoisser2016going} to select other points within  the model diameter $ d $ from the model to construct as point pairs.
	\subsubsection{Pose clustering}
	
	To merge similar candidates, we used a hierarchical clustering method \cite{vidal2018method}. If the rotation and transformation between the two candidate poses are less than the predefined threshold, the two candidate poses are grouped. All poses within each cluster follow the same conditions based on the two thresholds of rotation and transformation. Finally, the quaternion average for each cluster is used to calculate a new candidate pose, and the score of each pose is added up to the score of the new candidate pose.
	
	\begin{figure}[b!]
		\centering
		\includegraphics[width=1\columnwidth]{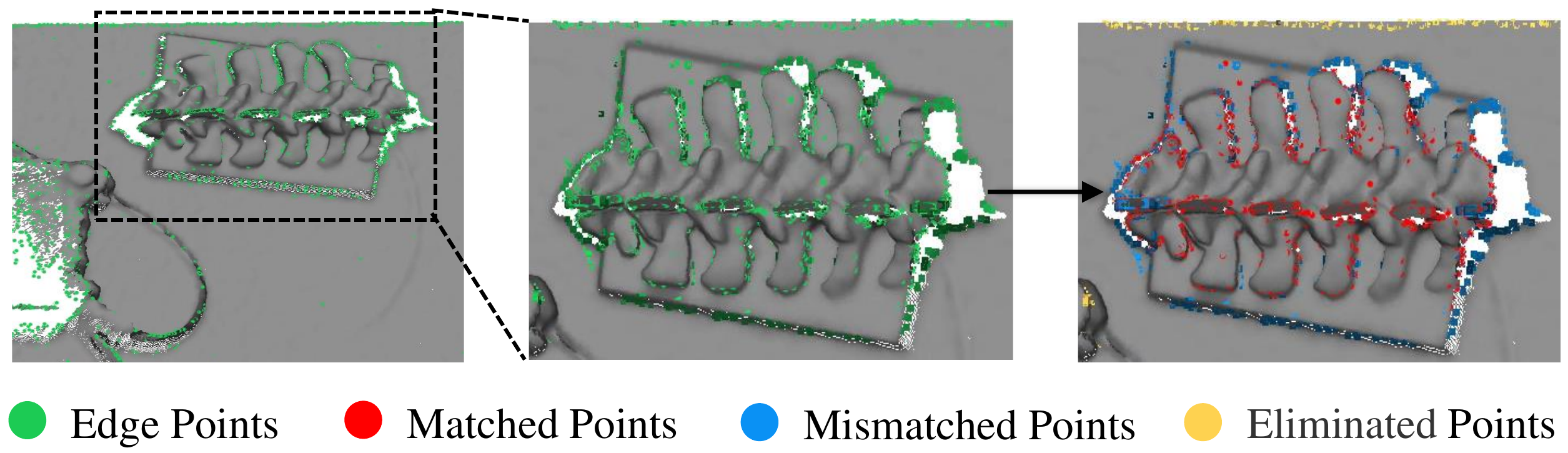}
		\caption{\label{fig:5} Classification of edge points in ROI.}
	\end{figure}
	
	\subsubsection{Post-processing}
	\label{sec:Postprocessing}
	The score of each pose is obtained by adding the votes of the candidates in the cluster. In the presence of sensor noise and background clutter, the score of the poses may not correctly represent the degree of matching.  Therefore, we recommend that a more reliable score be calculated through an additional re-scoring process. We observed that in most cases \cite{hinterstoisser2016going,vidal2018method,guo2021efficient, papazov2010efficient}, most of the computational time is spent on pose verification. So to ensure the time efficiency of pose estimation, we propose an edge-based pose hypothesis verification method with an early exit strategy.
	
	\par 
	Edges are distinctive features of an object and can strongly represent the shape characteristics and contours of the object. With the edge information of the point cloud, it is possible to select the correct pose from a set of candidate poses with high probability. In our pose hypothesis verification method, for the input candidate pose, the axis-aligned bounding box (AABB box) of the computed candidate pose is used as the ROI region. The edge points within the ROI region are clustered, and the distance between the edge clustering center and the center of the candidate pose is computed to remove remote and divergent edge points. The reason for using filtering based on the distance to the centroid of edge clustering is that often cluttered edges that are not in the object are discontinuous and distant. The final score for this candidate pose is shown in Eq. \ref{eq:2} below. $ N_{\text {ROI}} $ denotes the number of edge point clouds of ROI (red part and blue part in Fig.  \ref{fig:5}) after filtering out outliers (yellow part in Fig. \ref{fig:5}). $N_{\text {Matching}} $ is the number of edge point clouds close to the candidate poses (red part in Fig. \ref{fig:5}), and the degree of edge matching S is given by:
	\begin{equation}
		\text { S }=\frac{N_{\text {Matching }}}{N_{\text {ROI}}}.
		\label{eq:2}
	\end{equation}
	
	The specific steps of the pose verification process are as follows:
	\begin{itemize}
		
		\item 	The input candidate poses are sorted according to the number of votes, and the maximum number of votes for the candidate poses is ${V}_\mathrm{max} $. The candidate poses are divided into two categories according to $ {V}_\mathrm{max} $. The first one is the candidate pose with the number of votes greater than $ {V}_\mathrm{max}/2 $, which is more likely to be the correct pose. The second category is the candidates with less than $ {V}_\mathrm{max}/2 $. The number of candidates in this category is much larger than in the first category.

		\item In the first category of candidate poses, we use KD-tree to quickly see how well each pose matches the edges of the scene. Those edge points that are close to the model indicate support for the pose hypothesis, after which the $N$ candidate poses with the highest scores (the value of $N$ is given in Section \ref{sec:Parameteranalysis}) are selected for more detailed filtering using Eq. \ref{eq:2}. The reason why we do not directly use the edge match of the whole scene point is that the correctness of the match is greatly reduced when the scene is prone to clutter. If the pose score computed by Eq. \ref{eq:2} is higher than 0.7, it is directly considered as the correct pose and the subsequent computation is stopped. If the calculated $N$ poses are lower than 0.7 but higher than 0.6, the one with the highest score among the $N$ poses is selected.

		\item If the calculated score of the $N$ poses of the first category is lower than 0.6, the poses of the second category are processed in the same way as the poses of the first category above. If the $N$ poses of the second category are also not higher than 0.6, the pose with the highest score from the $2N$ candidate poses is selected as the final pose.
	\end{itemize}

\par	After selecting the final pose, ICP\cite{chen1992object} is used to further refine the pose to improve the accuracy of the match.
	
	\begin{figure}[t!]
		\centering
		\includegraphics[width=1\columnwidth]{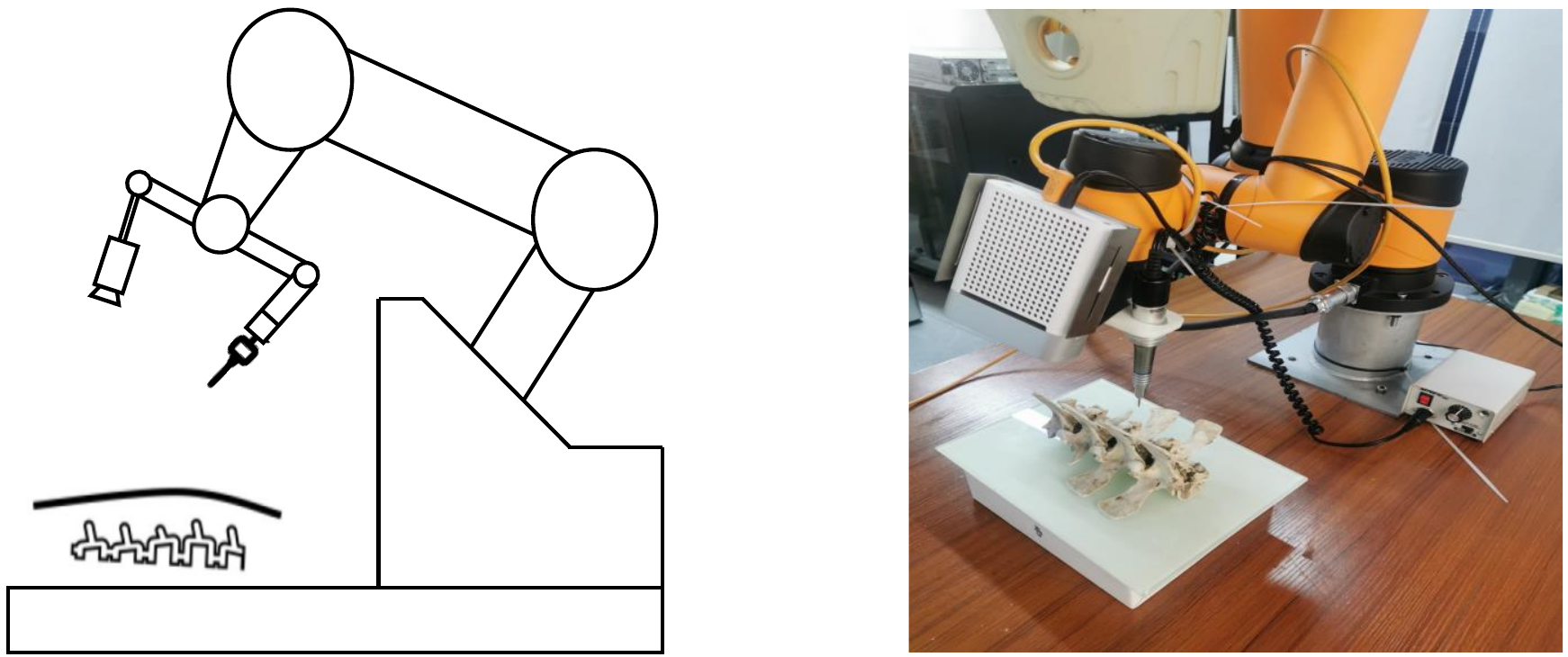}
		\caption{\label{fig:7} Hardware composition of our experiment. The left is the schematic diagram, the right is the physical diagram.}
	\end{figure}

	\subsection{Design of robot-operated positioning experiment}

	\label{Designofsurgicalnavigationsystem}
	\subsubsection{Hardware composition}

	The hardware composition of our experiment  is shown in Fig. \ref{fig:7}. The 3D camera used in the experiments is the Azure Kinect DK depth camera. The robotic arm is the AUBO collaborative robot with six joints for flexible operation, and it is used to perform fixed-point movements to complete operations on the spine. The medical drill is fixed at the end of the robotic arm and is equipped with various drill holes, adjusted for different speeds, pointing at the spine. We build the platform  not only in the real environment but also in the simulation environment.

	\subsubsection{Transformation relationship analysis}
	
	In order to control the drill mounted on the robotic arm to be able to drill in the attitude we specify, we perform the coordinate transformation. We should make it clear that the transformation relationship is between the model of the spine, the fixed drilling, the depth camera, the end of the robotic arm and the base of the robotic arm. Finally, we should obtain the expected conversion relationship between drilling and the base of the robotic arm.
	\par
	First, based on the preoperative surgeon's design, we can obtain the target drill pose and position under the spine model coordinate system in advance and notate it as $ T_{t_\mathrm{hope}}^{s}  $; After the hand-eye calibration process to get the matrix notated $ T_{c} ^{e} $ that converts the coordinate system of the camera to the coordinate system of the end of the robotic arm; After the tool calibration process to get the matrix notated $ T_{t}^{e}  $ that converts the coordinate system of the fixed drilling to the coordinate system of the end of the robotic arm; The transformation matrix from the spine model coordinate system to the camera coordinate system is obtained from the above pose estimation algorithm, denoted as $  T_{s}^{c} $; The end effector's pose in the robotic base's coordinate system can be retrieved through the robotic arm's controller, and the current pose is notated as $ T_{e_{0}}^{b} $. In the fixed drill coordinate system, the transformation relationship from a fixed drill to the expected drill attitude is :

\begin{equation}
	 T_{t_\mathrm{hope}}^{t}=T_{e}^{t} \cdot T_{c}^{e} \cdot T_{s}^{c} \cdot T_{t_\mathrm{hope}}^{s}.
\end{equation}
	Finally, the expected conversion relationship between the drilling and the robot arm base is:

	\begin{equation}
	T_{t_\mathrm{hope}}^{b}= T_{e_{0}}^{b} \cdot T_{t}^{e} \cdot T_{t_\mathrm{hope}}^{t} .
\end{equation}

	\subsubsection{Position-based visual servoing scheme}
	
	\begin{figure}[b!]
		\centering
		\includegraphics[width=1.0\columnwidth]{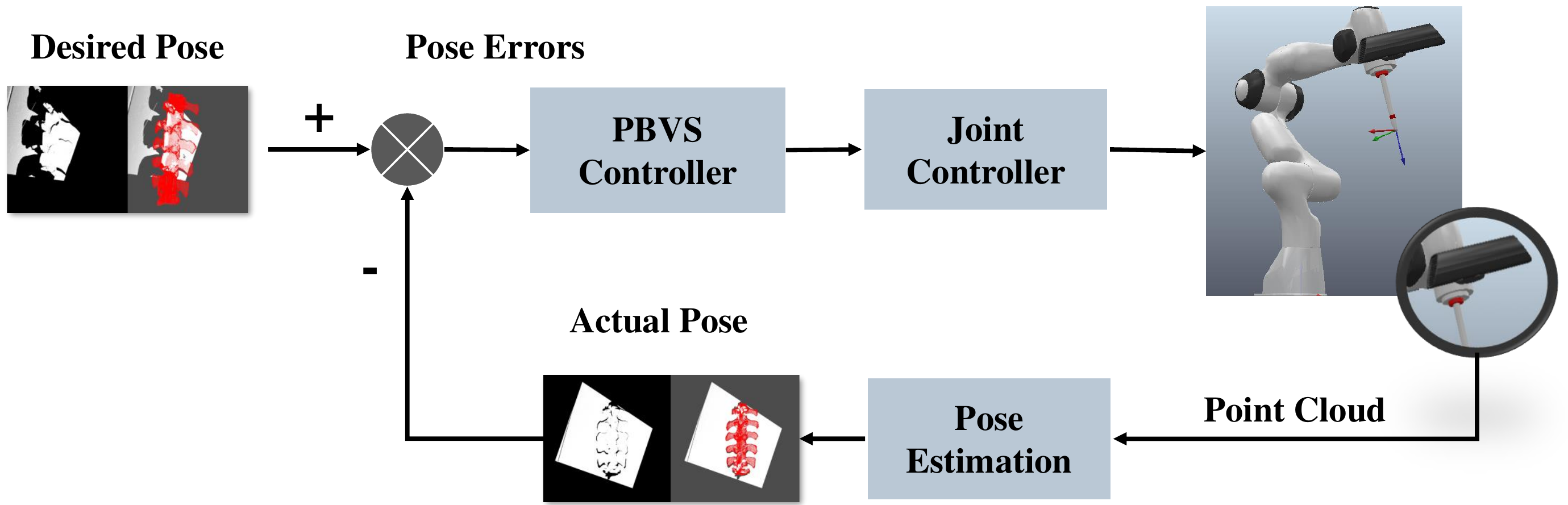}
		\caption{\label{fig:servoing}Position-based visual servoing scheme.}
	\end{figure}

Visual servoing uses visual information extracted from images or point clouds captured by one or more cameras to control the motion of a robot. Visual servoing is a closed-loop system in which vision analysis provides guidance for the robot and robot motion provides new vision analysis for the camera. Closed-loop design can effectively improve the success rate and reduce the deviation.

	We use a position-based visual servoing scheme, as shown in Fig. \ref{fig:servoing}. The input is the difference between the detected actual pose of the spine and the desired spine pose. The output is the control command of the robot velocity domain, and its purpose is to make the robot move quickly to the target pose state. After the instruction is completed, the camera continues to receive the feedback value of the robot state, forming a closed-loop control system. The closer the real pose is to the desired pose, the smaller the speed of the robot arm will be. When the difference is less than the threshold we set, the speed of the robot arm is 0, and the servo stops.

	\section{Experiments}
	\label{sec:Experiments}
	
	In this section, after describing the datasets required for the experiments, the evaluation criteria, and the state-of-the-art open-source comparison methods, we first evaluate the impact of different parameters on the real spine dataset. Then, in Sections \ref{sec:Qualityandrobustness} and 
	\ref{sec:Ablationstudy}, a real spine dataset and a publicly available dataset are tested together to investigate the robustness of the algorithm and the validity evaluation of algorithm design. In Section \ref{sec:Effectofsystem}, we evaluate our method quantitatively and qualitatively on the real spine dataset and show the result of the robot-operated positioning experiment. Finally, to demonstrate the effectiveness of our pose estimation method for objects with symmetry and complexity and its generality for objects of different shapes, we perform a comprehensive comparison of recognition rates and efficiency with state-of-the-art methods on two well-known publicly available datasets in Section \ref{sec:Effectofpublicdataset}.
	
	The algorithm proposed in this paper is implemented in the Point Cloud Library (PCL) and tested on a PC with a 3.6 GHz Intel(R) Core(TM) i9-10850K CPU and 16GB of RAM, and the algorithm uses OpenMP technology to improve the matching speed. 
	\subsection{The datasets}
	\label{sec:Thedataset}
	
	\subsubsection{The pubic datasets}
	
\begin{figure}[t!]
		\centering
		\includegraphics[width=1\columnwidth]{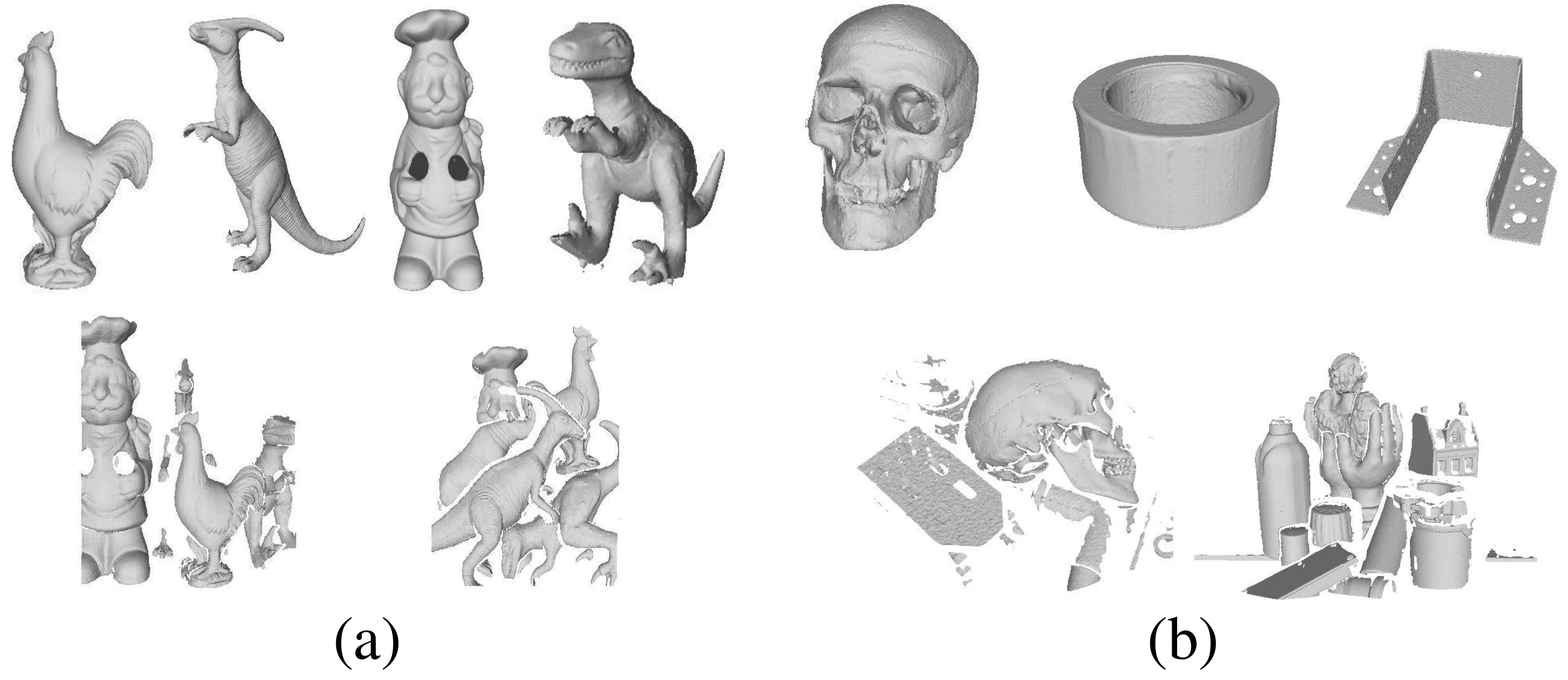}
		\caption{\label{fig:8} Several object models and two random scenes in the open datasets (a) UWA dataset; (b) DTU dataset.}
	\end{figure}

	The public dataset contains both UWA dataset \cite{mian2006three} and DTU dataset \cite{solund2016large}. The UWA dataset contains 5 complete 3D models as well as 50 2.5D scenes, where the rhino models are mainly used for interference. Each 2.5D scene contains four to five models, and the degree of model occlusion ranges from $ 65\% $ to $ 95\% $. 5 models and some scenes are shown in Fig. \ref{fig:8}(a). The DTU is a large dataset consisting of 45 objects and 3,204 scenarios captured by a structured light scanner, each of which contains 10 objects. These objects belong to three different types: geometrically complex models, cylindrical and planar models. Because some objects are highly occluded. We do not consider objects with more than $ 98\% $ occlusion. The DTU dataset is challenging because of the high occlusion, high similarity, and diversity of models. Some of the models and scenes are shown in Fig. \ref{fig:8}(b).
	
	\subsubsection{Spine dataset}
	
	\begin{figure}[t!]
		\centering
		\includegraphics[width=1\columnwidth]{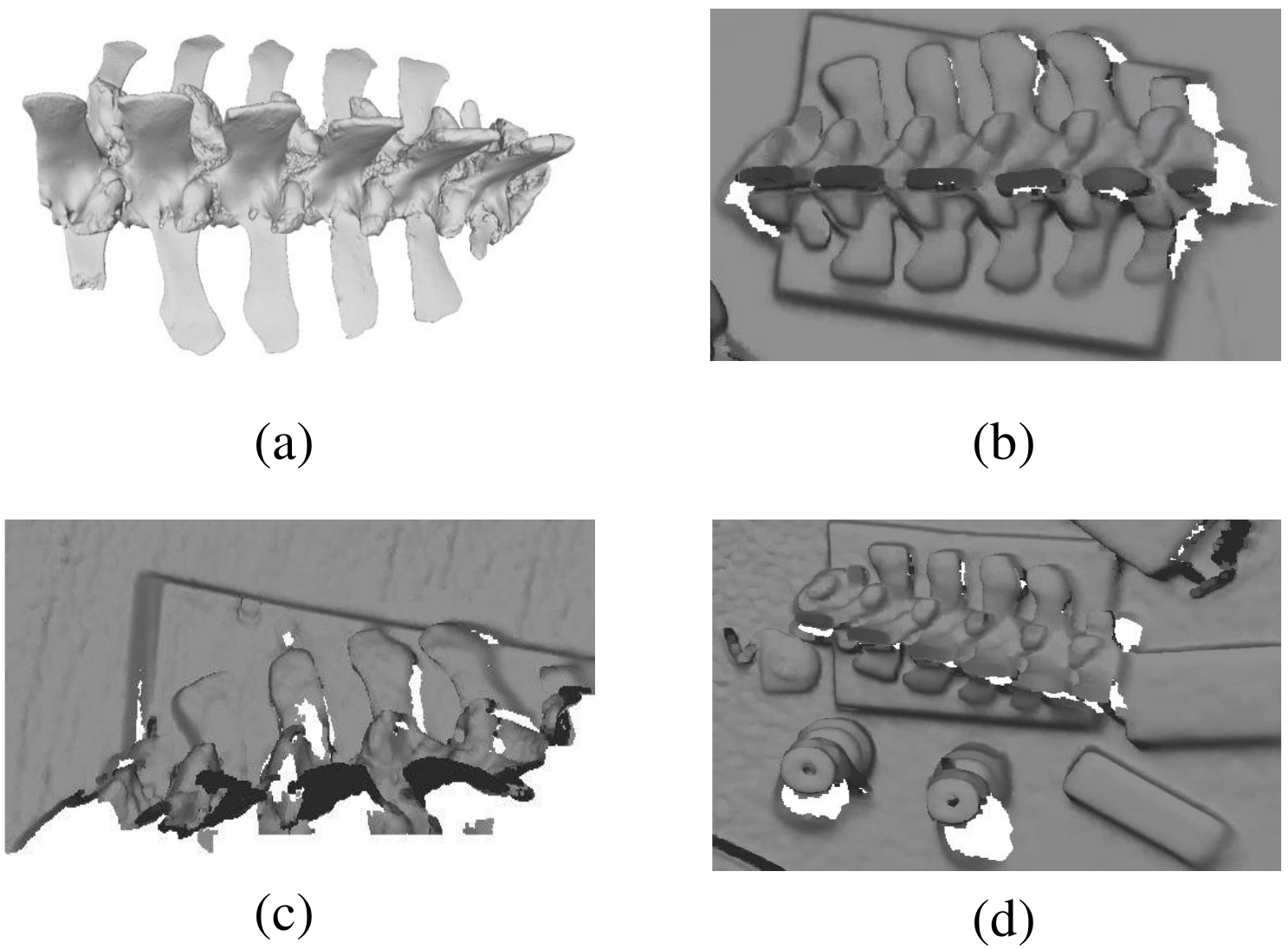}
		\caption{\label{fig:9} Spine dataset. (a) spine model, (b), (c), (d) S1, S2, S3, respectively.}
	\end{figure}
	To validate the effectiveness of our algorithm for spinal bone pose estimation, we construct a real dataset of the pig spine. The spine model point cloud uses CT scanning of the spine for accurate reconstruction, and then we use professional medical software Mimics Research to convert medical data in DICOM format into 3D models. The experimental platform  built in Section \ref{Designofsurgicalnavigationsystem} is used to collect three types of spine datasets with an Azure Kinect DK depth camera:
	\begin{itemize}
		\item less occluded far-field spine scenes [S1, Fig. \ref{fig:9}(b)];
		\item more occluded near-field spine scenes  [S2, Fig. \ref{fig:9}(c)];
		\item cluttered randomly placed spine scenes  [S3, Fig. \ref{fig:9}(d)];.
	\end{itemize}
	S1, S2, S3 have 40 scenes respectively, and the total number of datasets is 120.
	
	\begin{figure*}[t!]
		\centering
		\includegraphics[width=2.0\columnwidth]{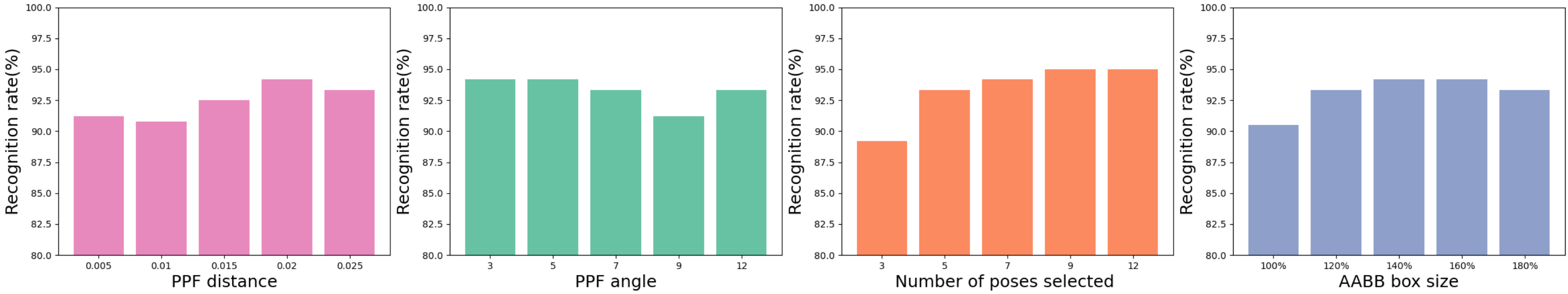}
		\caption{\label{fig:buchong_1} Parameter analysis for spine dataset. The default values of these parameters: the quantization step of distance $\Delta$$dist=0.025$ and the quantization step of angle $\Delta$$angle=5$, the number of poses using pose verification function $N=10$, and the size of AABB box $s=150\%$.}
	\end{figure*}
	
	\subsection{Evaluation criteria}
	\label{sec:Evaluationcriteria}
	
	To determine the pose accuracy, we adopt the Average Distance Metric (ADM) \cite{ hodavn2016evaluation} as the pose error metric. It considers both the visible and invisible parts of the 3D model surface. ADM measures the mean Euclidean distance between the model points converted by an estimated pose $ \hat{\mathbf{T}} $ and by the true pose $ \bar{\mathbf{T}} $, respectively. In  \cite{ guo2021efficient}, two alternatives of ADM (ADD and ADI) are used to define objects that do not have symmetric properties and those that do. We also use this evaluation criterion. And we accept the pose estimation as positive if the pose error is less than $\zeta_e$ , where $\zeta_e$  is related to the object diameter d. The pose error metrics of ADD and ADI are given by:
	\begin{align}
		\centering
		e_\mathrm{A D D} &=\underset{\mathbf{x} \in \mathcal{M}}{\operatorname{avg}}\|\overline{\mathbf{T}} \mathbf{x}-\hat{\mathbf{T}} \mathbf{x}\|_{2}, 
	\end{align}
	\begin{align}
		e_\mathrm{A D I}^{\prime} &=\text{max}(\underset{\mathbf{x}_{1} \in \mathcal{M}}{\operatorname{avg}} \min _{\mathbf{x}_{2} \in \mathcal{M}}\left\|\overline{\mathbf{T}} \mathbf{x}_{\mathbf{1}}-\hat{\mathbf{T}} \mathbf{x}_{\mathbf{2}}\right\|_{2},
		\left\|\overline{\mathbf{T}} \mathbf{c}_{o }-\hat{\mathbf{T}} \mathbf{c}_{o}\right\|_{2}),
	\end{align}
where $\mathcal{M}$ is the point cloud of model, $c_o$ is the object center. $e_\mathrm{ADD}$ is computing the average Euclidean distance of the same point after the transformation, while $e_\mathrm{ADI}$ is computing the average Euclidean distance of the two closest points  after the transformation and also takes into account the distance of the object center.

	\par 
		In this paper, we use two evaluation criteria Recognition Rate (RR) and Mean Recall (MR) to evaluate the performance of the algorithm. RR is the ratio of correct poses to all detected poses. MR is the average recognition rate of all objects and is used to measure the detection quality of the algorithm in the entire dataset:
		\begin{equation}
		M R=\operatorname{avg}_{o \in O} \frac{\sum_{s \in S}|P(o, s)|}{\sum_{s \in S}|G(o, s)|},
		\end{equation}
	
\noindent where O and S are the sets of all template objects and scenes, respectively, $P (o,s)$ is the set of correctly detected poses, and $G(o,s)$ is the set of  ground-truth poses of object $o $ in scene $s$.

	\subsection{Algorithms for comparison}
	\label{sec:comparisonofAlgorithms}
	We compare our method with several baselines using only depth images as input: Drost-PPF \cite{drost2010model}, Buch-17 \cite{ buch2017rotational}. We choose the commercial machine vision software MVTec HALCON to implement the original PPF and the optimization algorithm. 
	The open source method Buch-17 \cite {buch2017rotational} is a 3D object recognition method. It uses various three-dimensional local feature descriptors to find point pair correspondences that are constrained to vote in a 1-DoF rotation subgroup of the entire pose, SE (3). Kernel density estimation allows for an efficient combination of voting to determine the resulting pose. The method relies on three-dimensional local feature descriptors, which are evaluated with several descriptors, ECSAD \cite{jorgensen2015geometric}, NDHist\cite{buch2016local}, SI \cite{johnson1999using}, SHOT \cite{salti2014shot}, FPFH \cite{5152473}, and PPF \cite{drost2010model}.

	\begin{table}[!t]
		\centering
		\caption{ Results of our algorithm after interference by various noises}
		\scalebox{0.86}{
			\setlength{\tabcolsep}{0.5mm}
			\renewcommand\arraystretch{1.3}{
				%% \tablesize{} %% You can specify the fontsize here, e.g., \tablesize{\footnotesize}. If commented out \small will be used.
				\begin{tabular}{c|c|c|c|c|c|c}
					\toprule
					% \hline
					dataset  &  $\zeta_e$  & Noise=0  & Noise=0.5 & Noise=1.0 & Noise=1.5  & Noise=2.0

					\\\hline
					\multirow{2}{*}{Spine dataset}
					&  
					 $0.05d$    & 
					  94.17  & 
				89.17    &
				 85.83  & 
					80.83     & 
					78.33

					\\ \cline{2-7}
					
					\multirow{2}{*}{ }  &  $0.1d$ 
					
					& 95.00  
					& 90.00  
					&  87.50 
					& 83.33  
					& 80.83  
					\\ \hline
					\multirow{2}{*}{UWA dataset}
					&  $0.05d$  
					&  100.00
					&98.94 
					&  94.15
					& 89.36
					& 86.17 
					\\ \cline{2-7}
					
					\multirow{2}{*}{ }   
					&  $0.1d$
					& 100.00
					& 98.94
					& 94.15
					& 90.43
					& 86.70

					\\
					
					% \hline
					\bottomrule
		\end{tabular}}}
		\label{tab:noise}
	\end{table}
	
	\subsection{Parameter analysis}
	\label{sec:Parameteranalysis}
	In this subsection, we use the spine dataset for parametric analysis. To analyze each parameter, we use the variable control method for parameter validation. If the parameter does not have a determined value, we use the default value for the assignment. We mainly analyze the following four parameters: the quantization step of distance $\Delta$$dist$ and the quantization step of angle $\Delta$$angle$, the number of poses using pose verification function $N$, and the size of AABB box $s$. And the  $\Delta$$dist$  is related to the diameter of the Model. As shown in Fig. \ref{fig:buchong_1}, we can observe that the best performance is obtained with $\Delta$$angle=5$  and $\Delta$$distance=0.02$ , and the higher the number of selected poses, the higher the correct rate, but considering the time consumption, we set $N=9$. An axis-aligned bounding box(AABB) is the ROI used to calculate the pose verification function of candidate poses. The larger the size of the AABB box, the more points around the pose are considered, so it is easy to filter out some poses that only partially match the spine. We hope to determine the correctness of the poses by considering the matching degree of the points in the AABB box, but when the AABB box is larger than a certain degree, the accuracy of the poses is susceptible to the influence of outliers. The accuracy has a tendency to decrease, so we choose the AABB box size as 140\%.

	\subsection{Quality and robustness}
	\label{sec:Qualityandrobustness}
	In this subsection, we test the performance of our method in terms of Gaussian noise using the real bone dataset and the open dataset UWA. We randomly add Gaussian noise with different standard deviation values on the point coordinates. The standard deviations range from 0.0, 0.5, 1.0, 1.5, 2.0 (mm). Table \ref{tab:noise} shows the robustness of our method. The performance decreases slightly as the noise level increases, but we still perform well on the noisy data.

	\subsection{Ablation study}
	\label{sec:Ablationstudy}
	
	\subsubsection{Effect of sampling on performance}
	
		\begin{table}[!t]
		\centering
		\caption{\label{tab:1} Validation of edge-based sampling method.}
		\scalebox{0.77}{
			\setlength{\tabcolsep}{0.5mm}
			\renewcommand\arraystretch{1.5}{
				%% \tablesize{} %% You can specify the fontsize here, e.g., \tablesize{\footnotesize}. If commented out \small will be used.
				\begin{tabular}{c|c|c|c|c|c|c|c|c}
					\toprule
					% \hline
					Sampling Method  & 
					$\zeta_e$  &
					
					$ \mathrm{RR}_\mathrm{S1} $ & 
					$ \mathrm{RR}_\mathrm{S2} $ & 
					$ \mathrm{RR}_\mathrm{S3} $ & 
					$ \mathrm{RR}_\mathrm{chicken} $ & $ \mathrm{RR}_\mathrm{para} $ & $ \mathrm{RR}_\mathrm{cheff} $ & $ \mathrm{RR}_\mathrm{Trex} $
					
					\\\hline
					\multirow{2}{*}{Ref. \cite{guo2021efficient} }
					&  0.05d
					&  $\mathbf{92.5 }$
					&  $77.5 $
					&   $90.0 $ 
					&   $\mathbf{97.9 }$
					&   $97.8$ 
					&   $\mathbf{98.0 }$
					&   97.8
					 \\ \cline{2-9}
					 
					\multirow{2}{*}{ }  
					& $0.1d$ 
					& $\mathbf{92.5 }$
					& $80.0 $ 
					& $90.0 $ 
					&$97.9 $ 
					& $97.8$ 
					& $98.0$
					& $\mathbf{100.0 }$

					\\ 
					\hline
					\multirow{2}{*}{OURS }
					 &  0.05d
					 &  $\mathbf{92.5 }$ 
					 & $\mathbf{80.0 }$  
					 &  $\mathbf{92.5 }$  
					 &  $\mathbf{97.9 }$  
					 &  $\mathbf{100.0 }$ 
					 &  $\mathbf{98.0}$ 
					 &  $\mathbf{100.0 }$ 
					 \\ \cline{2-9}
					 
					 \multirow{2}{*}{ }  
					 & $0.1d$ 
					 & $\mathbf{92.5 }$ 
					 & $\mathbf{82.5 }$  
					 & $\mathbf{92.5 }$  
					 & $\mathbf{100.0 }$  
					 & $\mathbf{100.0 }$ 
					 & $\mathbf{100.0 }$ 
					 & $\mathbf{100.0 }$ 
					 
					  \\
					
					% \hline
					\bottomrule

		\end{tabular}}}
		
	\end{table}

	To clearly describe the contribution of the sampling in our method to the final result, we compare it with the sampling method \cite{ guo2021efficient} that does not emphasize the edge points. In order to make the number of points sampled by the method focusing on edge points smaller or equal to the compared method, we perform an additional sampling step for non-edge points. As shown in Table \ref{tab:1}, the result is a higher recall for sampling more focused on edge points, which we attribute to the fact that stable features are more present on the contours of the object. It has been shown that increasing the number of edge points sampled can improve the matching results.

	\subsubsection{Effect of pose verification function on performance}
	
	In this subsection, we test the edge-based post-processing method and pose verification method in  \cite{papazov2010efficient}. In  \cite{papazov2010efficient}, it is scored based on the overlap of surfaces, and those model points that are close to the scene vote to indicate support for the pose hypothesis. As shown in Table \ref{tab:2}, our edge-based post-processing approach is more discriminative. The edge information can robustly describe the geometric contour of the object. When in the ROI region, the higher the matching of edge points, the higher the probability that it is the correct pose.
	
	\begin{table}[!b]
		\centering
		\caption{\label{tab:2} Validation of our pose verification method.}
		\scalebox{0.77}{
			\setlength{\tabcolsep}{0.5mm}
			\renewcommand\arraystretch{1.5}{
				%% \tablesize{} %% You can specify the fontsize here, e.g., \tablesize{\footnotesize}. If commented out \small will be used.
				\begin{tabular}{c|c|c|c|c|c|c|c|c}
					\toprule
					% \hline
					Sampling Method  &
					$\zeta_e$ &
					$ \mathrm{RR}_\mathrm{S1} $ & 
					$ \mathrm{RR}_\mathrm{S2} $ & 
					$ \mathrm{RR}_\mathrm{S3} $ & 
					$ \mathrm{RR}_\mathrm{chicken} $ & $ \mathrm{RR}_\mathrm{para} $ & $ \mathrm{RR}_\mathrm{cheff} $ & $ \mathrm{RR}_\mathrm{Trex} $

					\\\hline
					\multirow{2}{*}{	Ref. \cite{papazov2010efficient} }
					
					&
					$0.05d$     & 
					$90.0 $ & 
					$82.5 $ & 
					$85.0 $ & 
					$91.7 $  & 
					$91.1 $ & 
					$98.0 $ &
					$95.6$
					\\ \cline{2-9}
					
					\multirow{2}{*}{ }  &
					$0.1d$     &
					 $92.5 $ & 
					 $82.5 $ & 
					 $85.0 $ & 
					 $93.8 $  & 
					 $91.1 $ & 
					 $98.0 $ &
					 $95.6$

						\\\hline
					\multirow{2}{*}{OURS }
					
					&
					$0.05d$     & $\mathbf{95.0 }$ &
					 $\mathbf{85.0 }$  &
					  $\mathbf{100.0 }$  &
					   $\mathbf{97.9 }$  &
					    $\mathbf{100.0 }$ &
					     $\mathbf{100.0 }$ &
					      $\mathbf{100.0 }$ 
					\\ \cline{2-9}
					
					\multirow{2}{*}{ }  &
					$0.1d$    & $\mathbf{97.5 }$ & $\mathbf{87.5 }$  & $\mathbf{100.0 }$  & $\mathbf{100.0 }$  & $\mathbf{100.0 }$ & $\mathbf{100.0 }$ & $\mathbf{100.0 }$

					  \\
					
					% \hline
					\bottomrule
		\end{tabular}}}
		
	\end{table}
	
	\begin{figure}[b!]
		\centering
		\includegraphics[width=1.0\columnwidth]{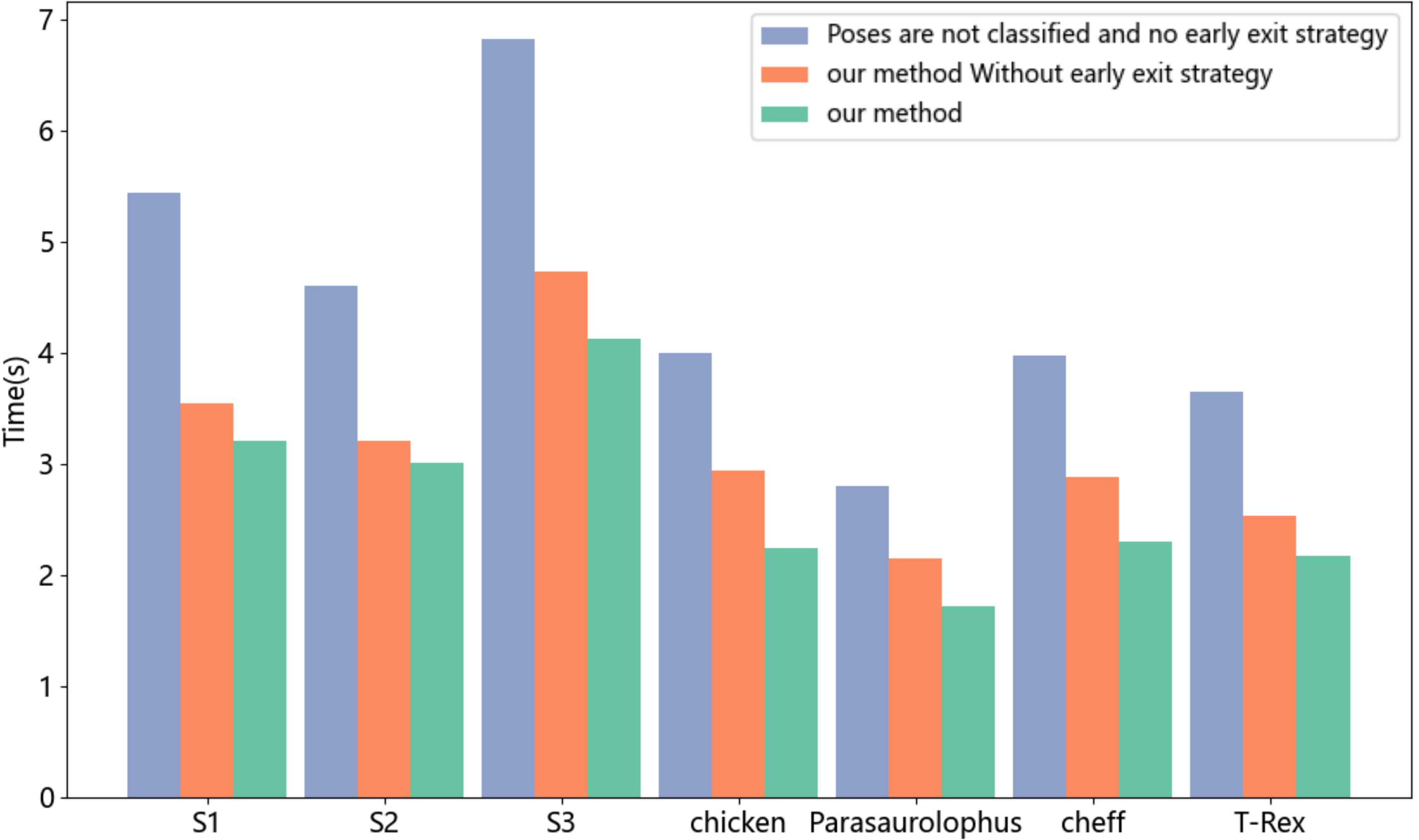}
		\caption{\label{fig:earlyexitstartegy} Comparison of time efficiency of three ways of using the pose verification functions based on UWA and spine datasets .}
	\end{figure}
	\subsubsection{Effect of Early exit strategy on performance}
	In this subsection, we focus on the time efficiency of our pose verification function, and we compare three ways of using the pose verification function. The first way is that the poses are not classified and then entered into the post-processing. The second way is as described in section \ref{sec:Postprocessing}, but without using an early exit strategy. The third way is our method in this paper, using the early exit strategy when the threshold is exceeded. As shown in Fig. \ref{fig:earlyexitstartegy}, the third one has the shortest time consumption. Our pose classification and an early-exit strategy have a greater improvement in efficiency. The reason why pose classification reduces time consumption is that poses with larger scores are more likely to be the correct pose. Therefore, processing firstly such category of poses with high likelihood and low number can reduce the time significantly.

	\begin{figure}[b!]
		\centering
		\includegraphics[width=1.0\columnwidth]{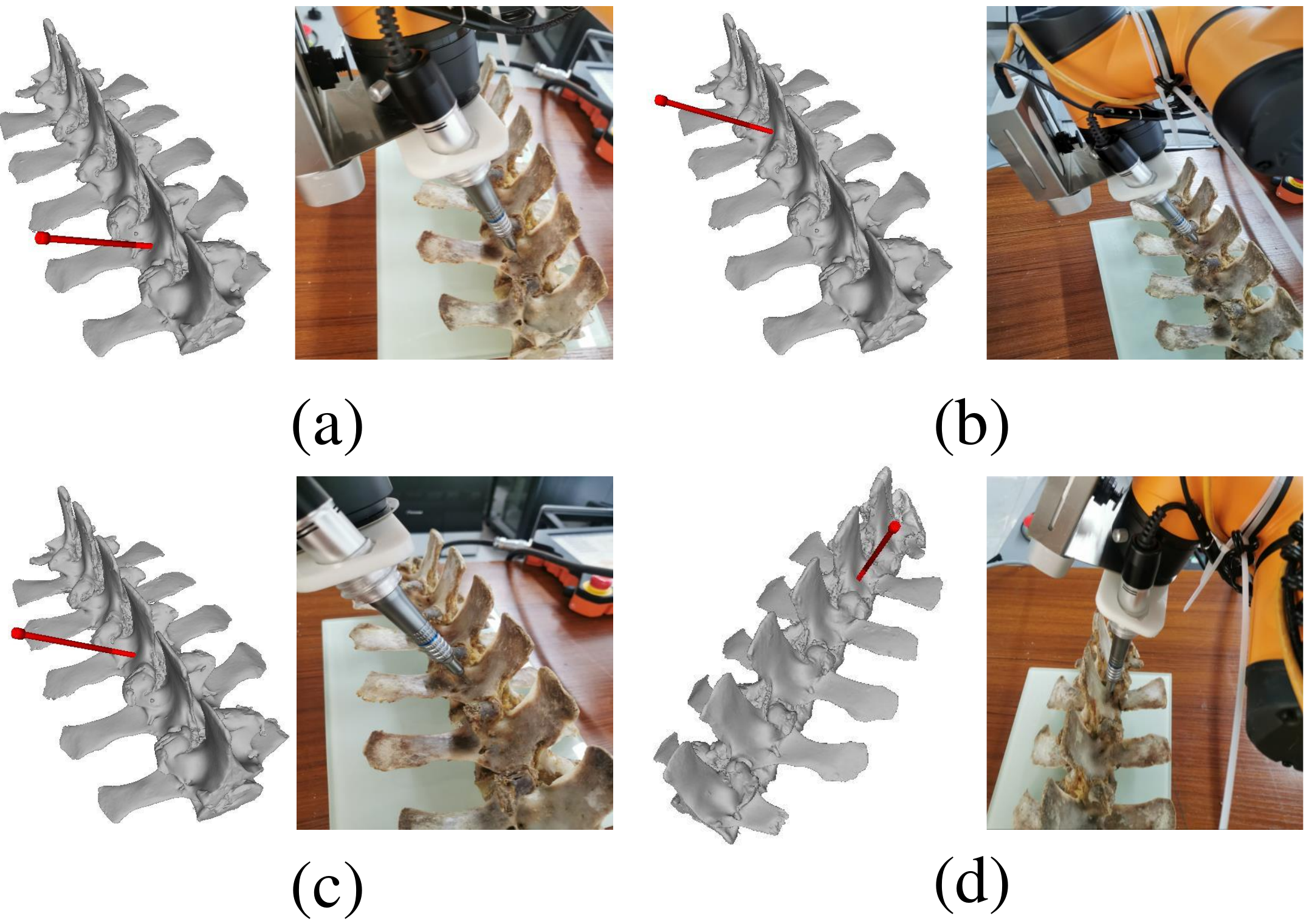}
		\caption{\label{fig:11} Predetermined pose and actual effect.}
	\end{figure}
	
	\subsection{Effect of the prototype system for operation}
	\label{sec:Effectofsystem}

	\begin{figure*}[t!]
		\centering
		\includegraphics[width=1.9\columnwidth]{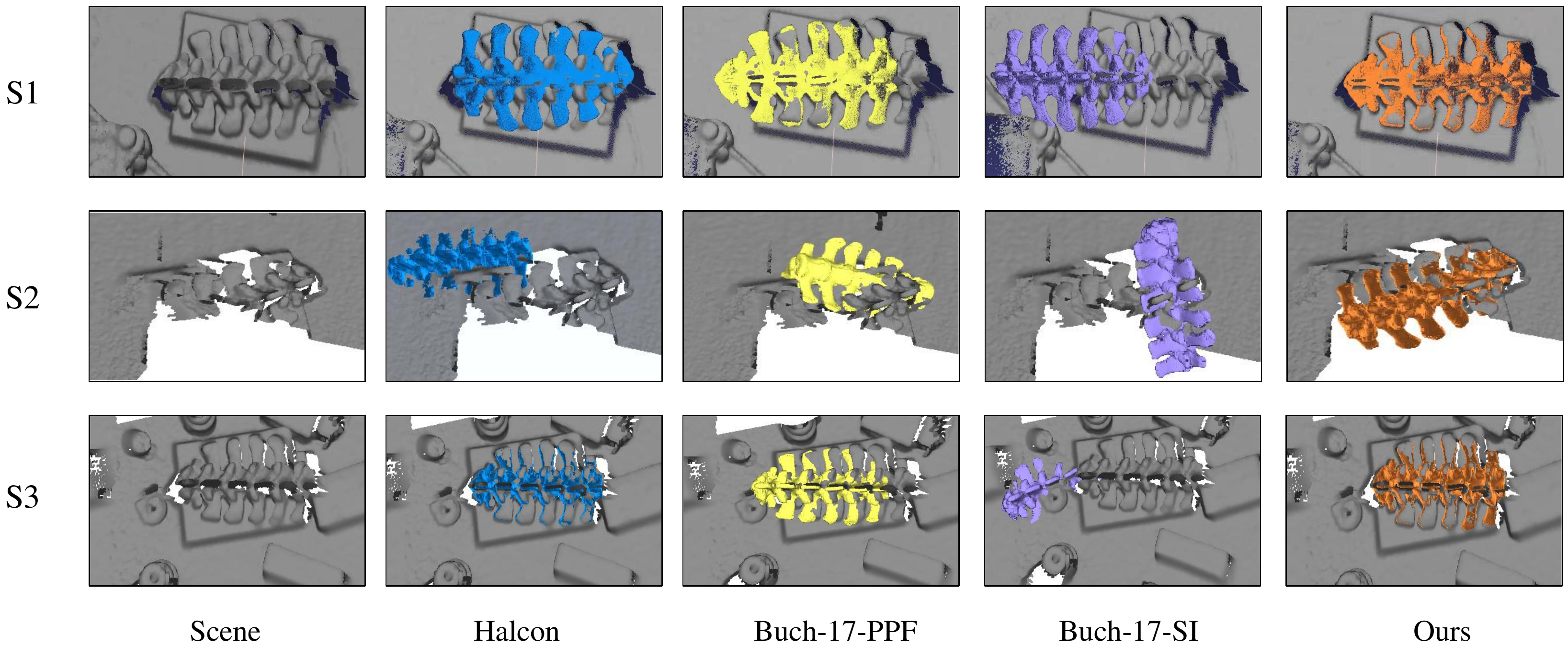}
		\caption{\label{fig:10} Qualitative comparison results on S1,S2,S3 scene from spine dataset.}
	\end{figure*}

	\begin{figure*}[t!]
		\centering
		\includegraphics[width=1.75\columnwidth]{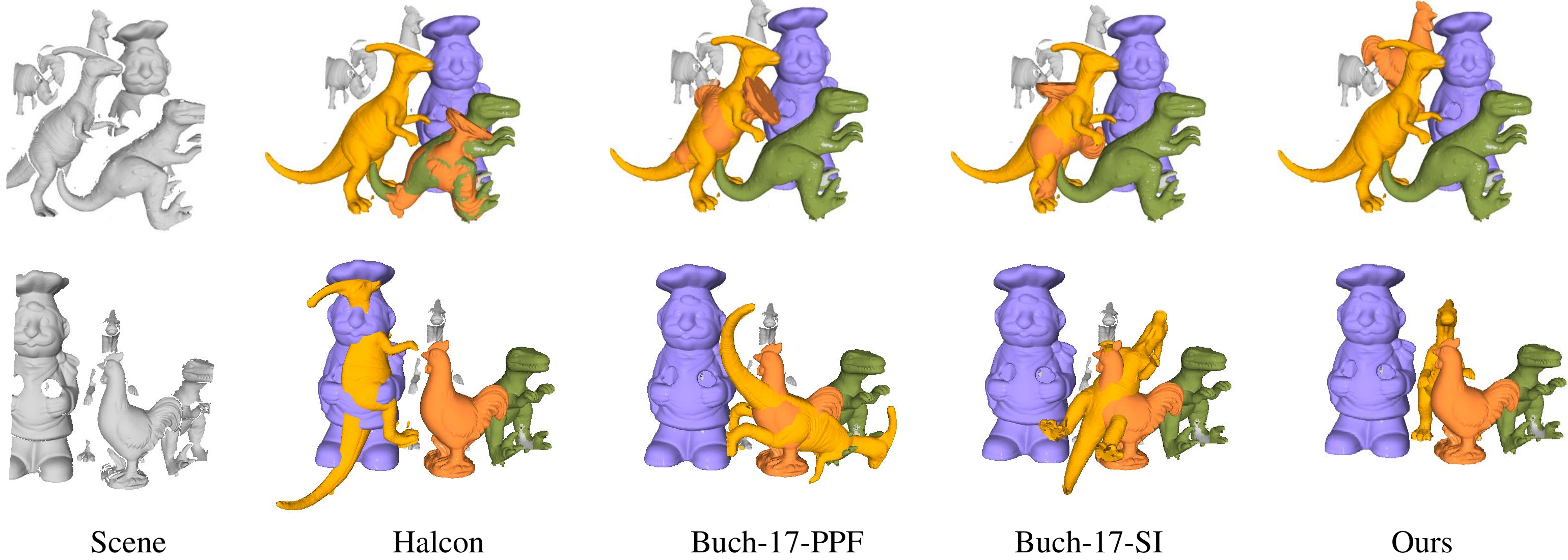}
		\caption{\label{fig:uwadataset} Qualitative comparison results on UWA dataset.}
	\end{figure*}
	\begin{figure*}[t!]
		\centering
		\includegraphics[width=1.75\columnwidth]{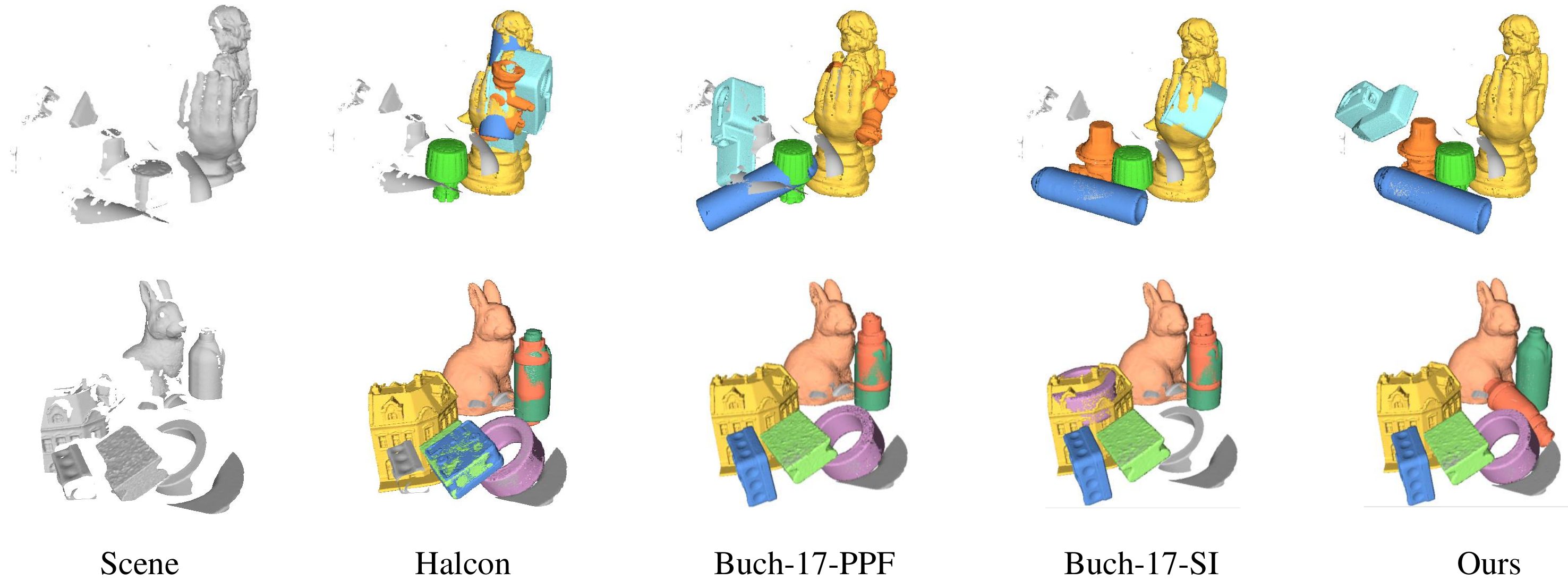}
		\caption{\label{fig:dtudataset1} Qualitative comparison results on DTU dataset.}
	\end{figure*}

	\begin{table}[!t]
		\centering
		\caption{\label{tab:3} Comparison of eight algorithms on the spine dataset.}
		\scalebox{0.84}{
			\setlength{\tabcolsep}{1.7mm}
			\renewcommand\arraystretch{1.2}{
				%% \tablesize{} %% You can specify the fontsize here, e.g., \tablesize{\footnotesize}. If commented out \small will be used.
				\begin{tabular}{c|c|c|c|c|c|c}
					\toprule
					% \hline
					Methods & 
					 $\zeta_e$  &
					$ \mathrm{RR}_\mathrm{S1} $ & 
					$ \mathrm{RR}_\mathrm{S2} $ & 
					$ \mathrm{RR}_\mathrm{S3} $ & 
					MR & Time
					
					\\\hline
					\multirow{2}{*}{Buch-17-ESCAD}
					
					&
					$0.05d$    &
					 $92.5 $ &
					  $77.5 $ &
					   $87.5 $ &
					    $85.8$ & 
					\multirow{2}{*}{$2.83 \mathrm{~s}$ }
					 
					\\ \cline{2-6}
					
					\multirow{2}{*}{ }  &
					$0.1d$    &
					 $95.0 $ & 
					 $77.5 $ &
					  $87.5 $ &
					   $86.7 $ &
					    \multirow{2}{*}{ }
					
						\\\hline
					\multirow{2}{*}{Buch-17-FPFH}
					
					&
					$0.05d$    &
					 $92.5 $ &
					  $70.0 $ & 
					  $77.5$ &
					   $80.0 $ & 
					\multirow{2}{*}{$8.76 \mathrm{~s}$ }
					
					\\ \cline{2-6}
					
					\multirow{2}{*}{ }  &
					$0.1d$    &
					 $92.5 $ &
					  $70.0 $ &
					   $77.5$ &
					    $80.0 $  &
					     \multirow{2}{*}{ }
					
					\\\hline
					\multirow{2}{*}{Buch-17-NDHIST}

					&
					$0.05d$   &
					 $95.0$ &
					  $77.5 $ &
					   $92.5 $&
					    $88.3$ & 
					\multirow{2}{*}{$3.28 \mathrm{~s}$  }
					
					\\ \cline{2-6}
					
					\multirow{2}{*}{ }  &
					$0.1d$    &
					 $95.0$ &
					  $80.0 $ &
					   $92.5 $ &
					    $89.2 $ & 
					     \multirow{2}{*}{ }

					\\\hline
					\multirow{2}{*}{Buch-17-PPF}
					
					&
					$0.05d$   &
					 $97.5$ &
					  $80.0 $ &
					   $90.0 $ &
					    $89.2 $ & 
					\multirow{2}{*}{$3.62 \mathrm{~s}$  }
					
					\\ \cline{2-6}
					
					\multirow{2}{*}{ }  &
					$0.1d$   &
					 $97.5$ &
					  $80.0 $ & 
					  $92.5 $ & 
					  $90.0 $ & 
					   \multirow{2}{*}{ }

					\\\hline
					\multirow{2}{*}{Buch-17-SHOT}
					
					&
					$0.05d$  &
					 $90.0 $ & 
					 $65.0 $ & 
					 $80.0$ & 
					 $78.3$ & 
					\multirow{2}{*}{$5.96 \mathrm{~s}$ }
					
					\\ \cline{2-6}
					
					\multirow{2}{*}{ }  &
					$0.1d$ &
					 $90.0 $ &
					  $65.0 $ &
					   $80.0$ & 
					   $78.3$ &
					     \multirow{2}{*}{ }

					\\\hline
					\multirow{2}{*}{Buch-17-SI}
					
					&
					$0.05d$  &
					 $92.5$ & 
					 $72.5 $ &
					  $87.5 $ &
					   $84.2$ & 
					\multirow{2}{*}{$3.11 \mathrm{~s}$  }
					
					\\ \cline{2-6}
					
					\multirow{2}{*}{ }  &
					$0.1d$ &
					 $92.5$ &
					  $72.5 $ &
					   $87.5 $ &
					    $84.2$ &
					      \multirow{2}{*}{ }

						\\\hline
					\multirow{2}{*}{HALCON}
					
					&
					$0.05d$   &
					 $\mathbf{97.5 }$ &
					  $82.5 $&
					   $\mathbf{100.0}$ &
					    $ \mathbf{93.3}$ & 
					\multirow{2}{*}{$\mathbf{1.78} \mathrm{~s}$  }
					
					\\ \cline{2-6}
					
					\multirow{2}{*}{ }  &
					$0.1d$  & 
					$\mathbf{97.5 }$ &
					 $82.5 $ &
					  $\mathbf{100.0}$ &
					   $93.3$ &
					      \multirow{2}{*}{ }

					\\\hline
					\multirow{2}{*}{OURS}
					
					&
					$0.05d$   & 
					$\mathbf{95.0 }$ & 
					$\mathbf{85.0 }$  & 
					$\mathbf{100.0}$ & 
					$\mathbf{93.3 }$  & 
					\multirow{2}{*}{$3.45\mathrm{~s}$  }
					
					\\ \cline{2-6}
					
					\multirow{2}{*}{ }  &
					$0.1d$  & 
					$\mathbf{97.5 }$ & $\mathbf{87.5 }$  & $\mathbf{100.0}$ & $\mathbf{95.0 }$&   \multirow{2}{*}{ }

					 \\
					
					% \hline
					\bottomrule
		\end{tabular}}}
		
	\end{table}
	
	\subsubsection{Recognition results on the spine dataset}
	
	As shown from Table \ref{tab:3}, the algorithm in this paper achieves great results in terms of correctness compared to other algorithms. The results show that our algorithm outperforms the other competitors. In terms of time cost, the commercial software HALCON is the fastest because it makes full use of the hardware and is also fully optimized at each step.  Compared with  \cite{buch2017rotational}, our method is faster than most of the 3D descriptor algorithms. Our algorithm can subsequently be further accelerated at each step on the GPU for surgical navigation applications.  Fig. \ref{fig:10} shows a qualitative comparison of these methods for several scenes.
	\subsubsection{Results of navigation and positioning}
	
\begin{figure}[t!]
		\centering
		\includegraphics[width=1.0\columnwidth]{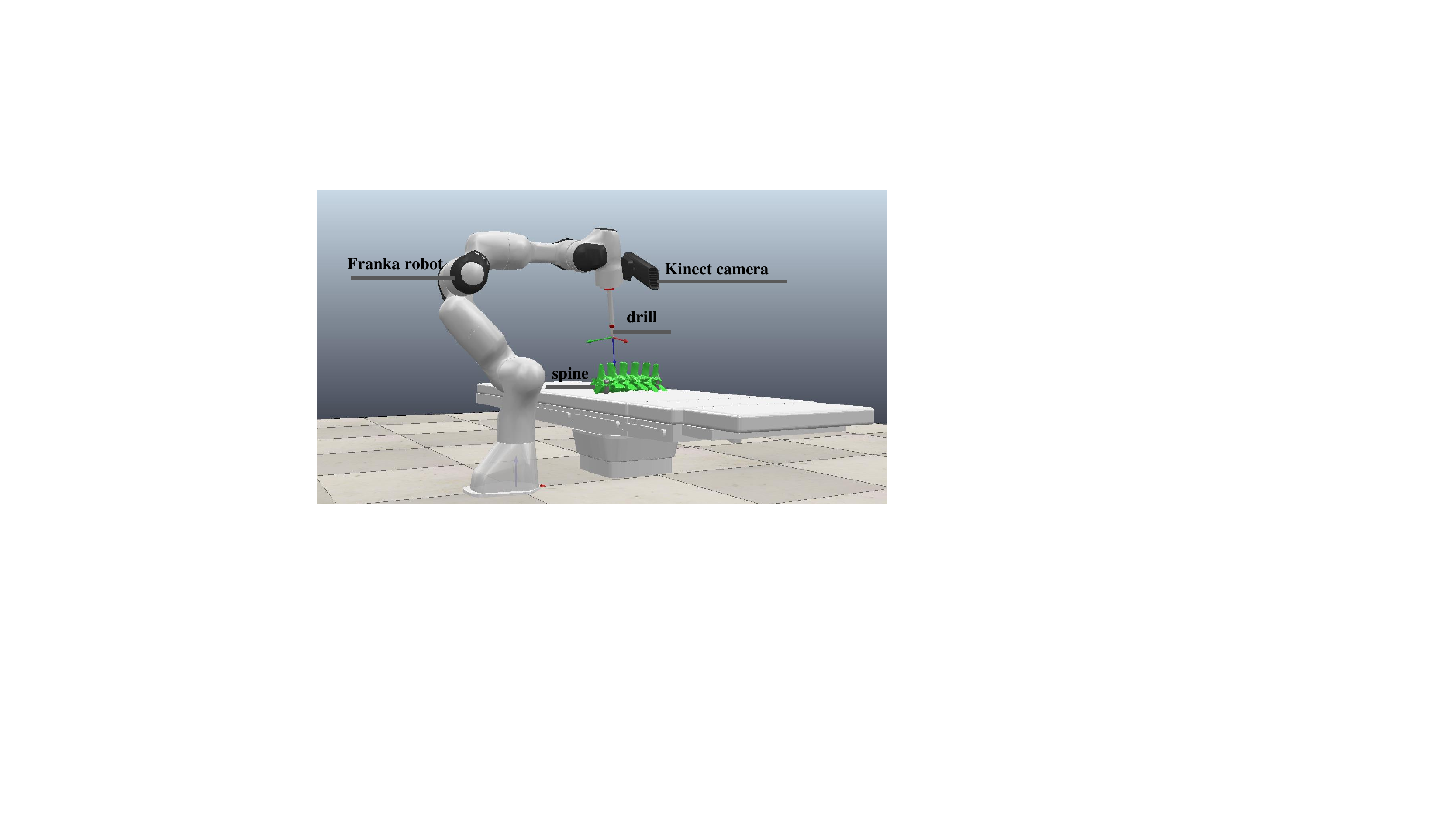}
		\caption{\label{fig:vrep} simulation environment. }
	\end{figure}

	In order to verify the effectiveness of the robot control method, we verify the feasibility of the scheme in the simulation environment. As shown in Fig. \ref{fig:vrep}, it shows the visualization interface, which is simulated in CoppeliaSim. In the simulation environment, camera intrinsics, hand-eye calibration parameters and tool calibration parameters can be directly calculated. However, in the real scene, these parameters can only be obtained by calibration, and there are errors in the calibration process, which can not be accurately calculated. To simulate the real situation, we add noise to these parameters. Based on some experience in real scenarios, add Gaussian noise of $\sigma$ = 5 for $f_x$, $f_y$ and $\sigma$ = 1 for $c_x$,$c_y$ in the camera intrinsics. Gaussian noise of $\sigma$ = 0.01 is added for the rotation and translation vectors of the calibration parameters.

\begin{figure}[t!]
			\centering
			\includegraphics[width=1.0\columnwidth]{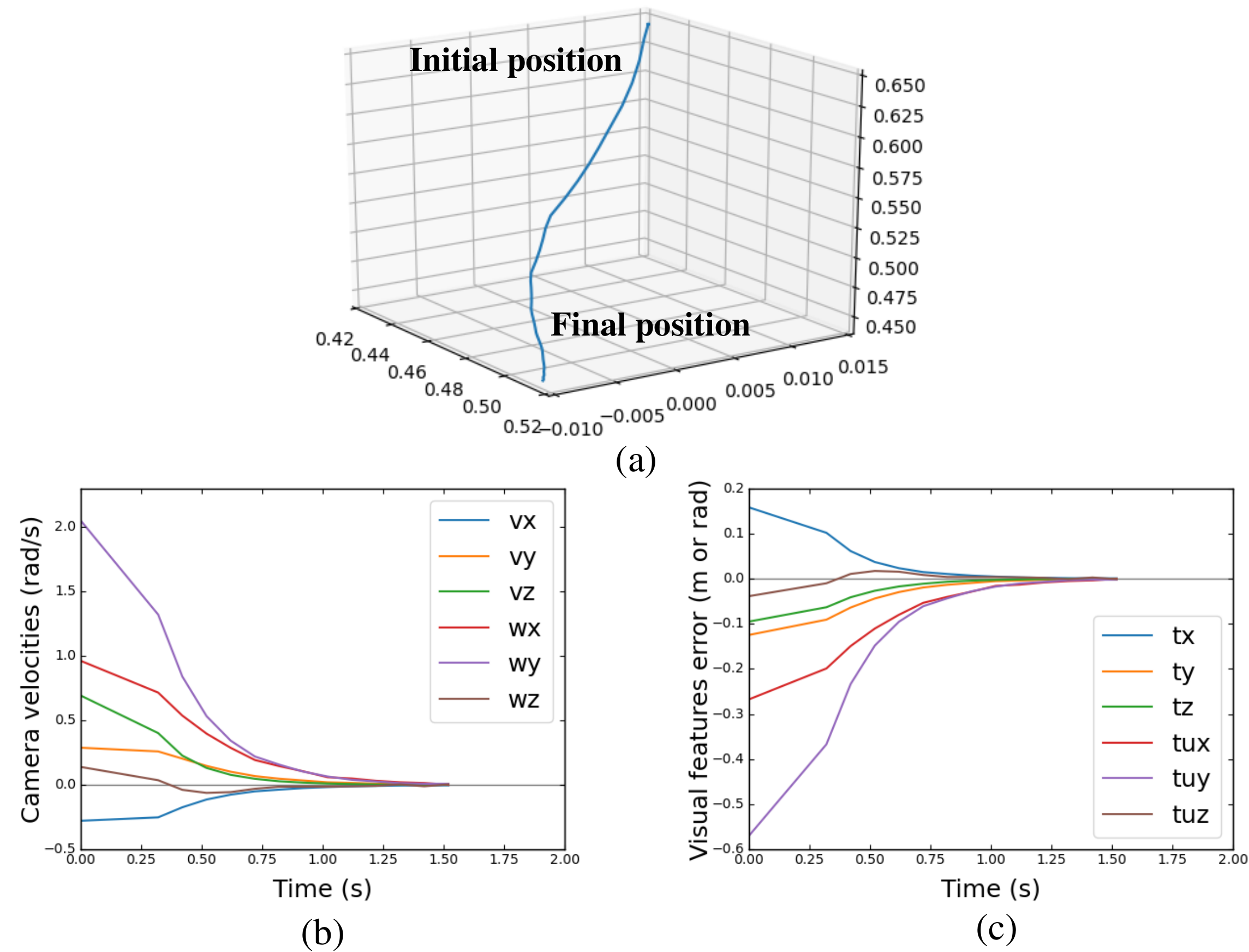}
			\caption{\label{fig:tracking}Experimental results of the simulation  (a) The motion trajectory of the camera's optical center  in Cartesian space. (b) Visual features error. (c) Camera  velocities. }
		\end{figure}

		Under this setting, the robot arm performs a movement of two seconds at a time. During the simulation, the motion trajectory of the camera's optical center (in Fig. \ref{fig:tracking}(a)), visual features error (in Fig. \ref{fig:tracking}(b)), and camera velocities (in Fig. \ref{fig:tracking}(c)) were recorded. It can be seen from the change of feature errors and camera speed that the closer the drill is to the target pose, the lower the speed of the robot arm. It is calculated that the tip distance error is within 1$mm$ and the angle error is within $1^\circ$.

	As shown in Fig.\ref{fig:11},  it shows the qualitative experimental results in the real environment. The left is a pose diagram of the prescribed drill, and the right is the robotic arm's effect.

	\begin{table*}[t!]
		\center
		\caption{\label{tab:DTU_SPINE}  Quantitative comparison of some complex, symmetric DTU models}
		\setlength{\tabcolsep}{0.4\tabcolsep}
		\renewcommand\arraystretch{1.2}{
			\begin{tabular}{c|c|c|c|c|c|c|c|c|c|c|c|c|c|c}
				\toprule
				Methods
				& $\zeta_e$
				& $ \mathrm{RR}_\mathrm{1} $ 
				& $ \mathrm{RR}_\mathrm{2} $ 
				& $ \mathrm{RR}_\mathrm{3} $ 
				& $ \mathrm{RR}_\mathrm{4} $  
				& $ \mathrm{RR}_\mathrm{5} $ 
				& $ \mathrm{RR}_\mathrm{7} $ 
				& $ \mathrm{RR}_\mathrm{26} $ 
				& $ \mathrm{RR}_\mathrm{36} $ 
				& $ \mathrm{RR}_\mathrm{37} $ 
				& $ \mathrm{RR}_\mathrm{38} $ 
				&  $ \mathrm{RR}_\mathrm{39} $ 
				&$ \mathrm{RR}_\mathrm{46} $
				& $ \mathrm{MR} $ 
			
				 \\
				\hline  
				\multirow{2}{*}{Buch-17-ECSAD}
				& $0.05d$ 
				& $ 41.61$ 
				&  $68.57$ 
				& $ 31.25$ 
				& $ 66.30 $  
				& $ 68.29 $ 
				& $ 91.59 $ 
				& $ 5.33 $ 
				& $ 4.00$ 
				& $ 5.08 $ 
				& $ 0.00 $ 
				&  $ 0.00 $ 
				& $ 4.44 $
				& $ 42.45 $ 
			\\	\cline{2-15}

				\multirow{2}{*}{ }
				& $0.1d$ 
				& $ 43.07$ 
				&  $68.57$ 
				& $ 31.25$ 
				& $ 67.93 $  
				& $ 69.51 $ 
				& $ 91.59 $ 
				& $ 5.33 $ 
				& $ 8.00$ 
				& $ 7.61 $ 
				& $ 0.00 $ 
				&  $ 0.00 $ 
				&$ 6.67 $
				& $ 43.04 $

				 \\
				\hline  
				\multirow{2}{*}{Buch-17-FPFH}
				& $0.05d$ 
				& $ 33.58 $ 
				& $ 40.71 $ 
				& $12.50 $ 
				& $54.35 $  
				& $ 69.51 $ 
				& $ 89.72$ 
				& $ 4.00 $ 
				& $ 8.00 $ 
				& $ 4.06 $ 
				& $ 0.00 $ 
				&  $ 0.00$ 
				& $ 0.00 $
				&  $ 34.28 $ 
				\\	\cline{2-15}

				\multirow{2}{*}{ }
				& $0.1d$ 
				& $ 33.58 $ 
				& $ 40.71 $ 
				& $12.50 $ 
				& $55.43 $  
				& $ 69.51 $ 
				& $ 90.65$ 
				& $ 5.33 $ 
				& $ 16.00 $ 
				& $ 5.58 $ 
				& $ 0.00 $ 
				&  $ 0.00$ 
				&$ 0.00 $
				& $ 34.68 $

				 \\
				\hline  
				\multirow{2}{*}{Buch-17-NDHIST}
				& $0.05d$ 
				& $53.28 $ 
				& $49.29$ 
				& $26.56 $ 
				&  $67.93$  
				&  $ 89.02 $ 
				&  $ \mathbf{93.46} $ 
				&  $ 20.00 $ 
				& $ 8.00 $ 
				& $8.12$ 
				& $ 0.00$ 
				&  $10.53 $ 
				& $ 35.56 $
				&  $ 45.96 $ 
				\\	\cline{2-15}

				\multirow{2}{*}{ }
				& $0.1d$
				& $53.28 $ 
				& $50.00 $ 
				& $26.56 $ 
				& $67.93$  
				& $ 89.02 $ 
				& $ \mathbf{93.46} $ 
				& $ 20.00 $ 
				& $ 12.00 $ 
				& $10.66$ 
				& $ 0.00$ 
				&  $10.53 $ 
				&$ 35.56 $
				& $ 46.27 $

				 \\
				\hline  
				\multirow{2}{*}{Buch-17-PPF}
				& $0.05d$ 
				&  $\mathbf{66.42}$ 
				&  $ 72.86$ 
				&  $39.06$ 
				&  $ 77.72 $  
				&  $93.90 $ 
				&  $ \mathbf{93.46} $ 
				& $34.67$ 
				& $16.00$ 
				& $ 14.21 $ 
				& $ 2.63 $ 
				&  $ 15.79$ 
				&$ 60.00 $
				& $ 57.05$ 
				\\	\cline{2-15}

				\multirow{2}{*}{ }
				& $0.1d$ 
				& $\mathbf{67.15}$ 
				& $ 75.00 $ 
				& $40.63$ 
				& $ 79.89 $  
				& $93.90 $ 
				& $ \mathbf{93.46} $ 
				& $ 36.00 $ 
				& $20.00 $ 
				& $ 16.24 $ 
				& $ 2.63 $ 
				&  $ 15.79$ 
				&$ 62.22 $
				& $ 57.77$

				 \\
				\hline  
				\multirow{2}{*}{Buch-17-SHOT}
				& $0.05d$ 
				& $33.58 $ 
				& $50.71$ 
				& $18.75 $ 
				& $ 57.07 $  
				& $ 57.76 $ 
				& $ 76.64 $ 
				& $5.33 $ 
				& $4.00$ 
				& $ 6.09$ 
				& $ 2.63$ 
				&  $ 0.00 $ 
				&$ 4.44 $
				& $ 35.35 $ 
				\\	\cline{2-15}

				\multirow{2}{*}{ }
				& $0.1d$ 
				& $34.31 $ 
				& $52.14 $ 
				& $20.31 $ 
				& $ 58.15 $  
				& $ 57.76 $ 
				& $ 76.64 $ 
				& $ 9.33 $ 
				& $ 12.00$ 
				& $ 10.15 $ 
				& $ 2.63$ 
				&  $ 0.00 $ 
				&$ 8.89 $
				& $ 36.48 $ 
				
				 \\
				\hline  
				\multirow{2}{*}{Buch-17-SI}
				&$0.05d$ 
				& $ 64.23 $ 
				& $ 63.57$ 
				& $37.50 $ 
				& $ 72.83 $  
				& $86.59 $ 
				& $91.59$ 
				& $20.00 $ 
				& $ 16.00 $ 
				& $ 10.15 $ 
				& $ 0.00 $ 
				&  $0.00 $ 
				& $ 15.56 $
				&  $ 50.49 $ 
				\\	\cline{2-15}

				\multirow{2}{*}{ }
				&$0.1d$ 
				& $ 64.23 $ 
				& $ 65.71 $ 
				& $40.63 $ 
				& $ 72.83 $  
				& $86.59 $ 
				& $ 92.52$ 
				& $24.00 $ 
				& $ 20.00 $ 
				& $ 16.75 $ 
				& $ 2.63 $ 
				&  $0.00 $ 
				&$ 15.56 $
				& $ 51.57 $

				 \\
				\hline  
				\multirow{2}{*}{HALCON}
				&$0.05d$ 
				& $ 48.91 $ 
				& $ 67.86$ 
				& $ 40.63 $ 
				& $ 68.48 $  
				& $81.71 $ 
				& $ 90.65 $ 
				& $ 61.33$ 
				& $20.00$ 
				& $ \mathbf{22.34} $ 
				& $ 42.11 $ 
				&  $ 10.53 $ 
				&51.11
				& $56.24 $ 
				\\	\cline{2-15}

				\multirow{2}{*}{ }
				&$0.1d$ 
				& $ 48.91 $ 
				& $ 67.86$ 
				& $ 42.19 $ 
				& $ 69.48 $  
				& $81.71 $ 
				& $ 91.59 $ 
				& $ 61.33$ 
				& $ \mathbf{24.00}$ 
				& $ \mathbf{26.90} $ 
				& $ 47.37 $ 
				&  $ 10.53 $ 
				&$ \mathbf{73.33}$
				& $ 57.32 $

				 \\
				\hline  
				\multirow{2}{*}{OURS}
				& $0.05d$ 
				& $60.58$
				& $ \mathbf{79.29} $ 
				& $  \mathbf{64.06} $ 
				& $   \mathbf{89.67} $ 
				& $    \mathbf{95.12}  $ 
				& $91.59 $ 
				& $ \mathbf{70.67} $
				& $ \mathbf{24.00} $ 
				& $ 20.30 $ 
				& $ \mathbf{60.53} $ 
				&  $ \mathbf{36.84} $ 
				&$ \mathbf{66.67} $
				& $ \mathbf{66.62}  $ 
				\\	\cline{2-15}

				\multirow{2}{*}{ }
				&$0.1d$ 
				& $60.58$
				& $ \mathbf{79.29} $ 
				& $  \mathbf{64.06} $ 
				& $   \mathbf{90.76} $  
				& $    \mathbf{95.12}  $ 
				& $92.52 $ 
				& $ \mathbf{78.67} $ 
				& $ \mathbf{24.00} $ 
				& $ 20.30 $ 
				& $ \mathbf{60.53} $ 
				&  $ \mathbf{42.11} $ 
				&$ \mathbf{73.33} $
				& $ \mathbf{67.21}  $

				\\

				\bottomrule
				
		\end{tabular}}
		\label{tab1}
	\end{table*}

	\begin{table}[!t]
	\centering
	\caption{\label{tab:4}  Comparison of eight algorithms on the UWA dataset.}
	\scalebox{0.75}{
		\setlength{\tabcolsep}{0.8mm}
		\renewcommand\arraystretch{1.5}{
			%% \tablesize{} %% You can specify the fontsize here, e.g., \tablesize{\footnotesize}. If commented out \small will be used.
			\begin{tabular}{c|c|c|c|c|c|c|c}
				\toprule
				% \hline
				Methods &
				$\zeta_e $ &
				 $ \mathrm{RR}_\mathrm{chicken} $ & $ \mathrm{RR}_\mathrm{para} $ & $ \mathrm{RR}_\mathrm{cheff} $ & $ \mathrm{RR}_\mathrm{Trex} $ & $ \mathrm{MR} $ &  Time  \\
				\hline  
				\multirow{2}{*}{Buch-17-ECSAD}&
				$0.05d$ &
				 $93.75 $&
				  $80.00 $ &
				   $\mathbf{100.00 }$ &
				    $80.00$ &
				     $88.83 $ &
				      \multirow{2}{*}{ $3.08 \mathrm{~s}$}  \\ \cline{2-7}
				\multirow{2}{*}{ }&
				$0.1d$ &
				 $93.75 $ &
				  $80.00 $ &
				   $\mathbf{100.00 }$ &
				    $80.00$ &
				     $88.83 $ &
				      \multirow{2}{*}{ } \\

				\hline
				\multirow{2}{*}{ Buch-17-FPFH}&
				$0.05d$ &
				 $91.67 $ &
				  $84.44$ &
				   $88.00 $ &
				    $86.67 $ &
				     $87.77 $ &
				        \multirow{2}{*}{ $6.12 \mathrm{~s}$}  \\ \cline{2-7}
				        
				\multirow{2}{*}{ }&
				$0.1d$ &
				 $91.67 $ &
				  $84.44$ &
				   $92.00 $ &
				    $88.89 $ &
				     $89.36 $  &
				      \multirow{2}{*}{ } \\

				\hline  
				\multirow{2}{*}{Buch-17-NDHIST}&
				$0.05d$&
				 $93.75$ &
				  $95.56 $ &
				   $\mathbf{100.00 }$ &
				    $\mathbf{100.00 }$ &
				     $97.34 $ & 
				     \multirow{2}{*}{ $3.61 \mathrm{~s}$}  \\ \cline{2-7}
				\multirow{2}{*}{ }&
				$0.1d$ & 
				 $93.75$ &
				  $95.56 $ &
				   $\mathbf{100.00 }$ &
				    $\mathbf{100.00 }$ &
				     $97.34 $ &
				      \multirow{2}{*}{ } \\

				\hline  
				\multirow{2}{*}{Buch-17-PPF}&
				$0.05d$ & 
				$93.75 $ & 
				$95.56 $ & 
				$\mathbf{100.00 }$ &
				 $\mathbf{100.00 }$ &
				  $97.34 $ &
				   \multirow{2}{*}{ $3.77 \mathrm{~s}$}  
				   \\ \cline{2-7}
				\multirow{2}{*}{ }&
				$0.1d$& 
				$93.75 $ &
				 $95.56 $ &
				  $\mathbf{100.00 }$ &
				   $\mathbf{100.00 }$ & 
				   $97.34 $  &
				    \multirow{2}{*}{ } \\

				\hline  
				\multirow{2}{*}{Buch-17-SHOT}&
				$0.05d$&
				 $89.58 $ &
				  $75.56 $ &
				   $98.00 $ &
				    $71.11 $ &
				     $84.04 $ &
				      \multirow{2}{*}{ $3.95 \mathrm{~s}$}  \\ \cline{2-7}
				\multirow{2}{*}{ }&
				$0.1d$ &
				 $89.58 $ &
				  $75.56 $ &
				   $98.00 $ &
				    $71.11 $ &
				     $84.04 $ &
				      \multirow{2}{*}{ } \\

				\hline  
				\multirow{2}{*}{Buch-17-SI}&
				$0.05d$  & 
				$93.75 $ & 
				$91.11 $ & 
				$\mathbf{100.00 }$ &
				 $97.78 $ & 
				 $95.74 $ & 
				 \multirow{2}{*}{ $4.12 \mathrm{~s}$}  \\ \cline{2-7}
				
				\multirow{2}{*}{ } &
				$0.1d$ &
				 $93.75 $ & 
				 $91.11 $ & 
				 $\mathbf{100.00 }$ &
				  $97.78 $ &
				   $95.74 $ & 
				   \multirow{2}{*}{ } \\

				\hline  
				\multirow{2}{*}{HALCON}&
				$0.05d$&
				 $91.67 $ &
				  $91.11$ &
				   $98.00 $ &
				    $95.56 $ &
				     $94.15 $ &
				      \multirow{2}{*}{ $0.4 \mathrm{~s}$}  \\ \cline{2-7}
				\multirow{2}{*}{ }&
				$0.1d$ &
				 $91.67 $ &
				  $91.11$ &
				   $98.00 $ &
				    $97.88 $ &
				     $94.68 $ &
				      \multirow{2}{*}{ } \\ 
				
				\hline  
				\multirow{2}{*}{OURS}&
				$0.05d$ &
				 $\mathbf{97.92 }$ &
				  $\mathbf{100.00 }$ & 
				  $\mathbf{100.00 }$ & 
				  $\mathbf{100.00 }$ & 
				  $\mathbf{99.47 }$ &
				  \multirow{2}{*}{ $2.11 \mathrm{~s}$}  \\ \cline{2-7}
				\multirow{2}{*}{ }&
				$0.1d$& 
				$\mathbf{100.00 }$ &
				 $\mathbf{100.00 }$ &
				  $\mathbf{100.00 }$ &
				   $\mathbf{100.00 }$ &
				    $\mathbf{100.00 }$ &
				     \multirow{2}{*}{ } \\

				% \hline
				\bottomrule
	\end{tabular}}}
	
\end{table}

	\begin{table}[t!]
		\centering
		\caption{\label{tab:6}  Comparison of eight algorithms on the DTU dataset.}
		\scalebox{0.9}{
			\setlength{\tabcolsep}{0.9mm}
			\renewcommand\arraystretch{1.3}{
				%% \tablesize{} %% You can specify the fontsize here, e.g., \tablesize{\footnotesize}. If commented out \small will be used.
				\begin{tabular}{c|c|c|c|c|c}
				
					\hline
					Methods &
					$\zeta_e $ &
					 \makecell{Geometrically  \\  complex }  &  Cylindrical & Planar & MR \\
					
					\hline  
					\multirow{2}{*}{Buch-17-ECSAD }
					&
					$0.05d$  &
					 $57.12 $ &
					  $33.35 $ &
					   $35.69 $&
					    $\mathrm{40.63}$
					     \\ \cline{2-6}
					\multirow{2}{*}{ }&
					$0.1d$ &
					 $57.83 $ &
					  $36.24 $ &
					   $38.23 $ &
					    $\mathrm{42.83}$ \\

					\hline  
					\multirow{2}{*}{Buch-17-FPFH }
					&$0.05d$ 
					 & $45.25 $
					  & $34.62 $ 
					  & $15.76$ 
					  & $\mathrm{33.98}$
					   \\ \cline{2-6}
					\multirow{2}{*}{ }&
					$0.1d$&
					 $45.88 $ &
					  $38.26 $ &
					   $17.61 $ &
					    $\mathrm{36.40}$ \\

					\hline  
					\multirow{2}{*}{Buch-17-NDHIST }
					& $0.05d$ &
					 $61.71$ &
					  $41.11$ &
					   $47.97$ &
					    $\mathrm{48.37}$ 
					    \\ \cline{2-6}
					\multirow{2}{*}{ }&
					$0.1d$ & 
					$61.86 $ & 
					$43.61$ & 
					$50.28 $ & 
					$\mathrm{50.16}$ \\

					\hline  
					\multirow{2}{*}{Buch-17-PPF }
					& $0.05d$   &
					 $71.60 $ & 
					 $51.95 $ & 
					 $61.88 $ & 
					 $\mathrm{59.53}$ 
					  \\ \cline{2-6}
					\multirow{2}{*}{ }&
					$0.1d$ &
					 $72.31 $ &
					  $54.05 $ & 
					  $64.58 $ & 
						  $\mathrm{61.37}$ \\

					\hline  
					\multirow{2}{*}{Buch-17-SHOT }
					& $0.05d$ &
					 $48.50 $ &
					  $27.07$ &
					   $36.85 $ &
					    $\mathrm{35.13}$ 
					     \\ \cline{2-6}
					\multirow{2}{*}{ }&
					$0.1d$ &
					 $49.36 $ &
					  $30.23$ & 
					  $39.04 $ & 
					  $\mathrm{37.45}$  \\ 
					
					\hline  
					\multirow{2}{*}{Buch-17-SI }
					& $0.05d$ &
					 $61.95 $  &
					  $38.96 $ & 
					  $27.00 $ & 
					  $\mathrm{43.21}$ 
					  \\ \cline{2-6}
					\multirow{2}{*}{ }&
					$0.1d$ &
					 $62.73 $ &
					  $42.23 $ &
					   $28.89 $ & 
					   $\mathrm{45.73}$ \\

					\hline  
					\multirow{2}{*}{HALCON }
					& $0.05d$ &
					 $69.38 $  &
					  $45.90 $  &
					   $55.62 $  &
					    $\mathrm{54.54}$ 
					     \\ \cline{2-6}
					\multirow{2}{*}{ }&
					$0.1d$ &
					 $70.49 $ &
					  $49.45 $ &
					   $56.89 $ &
					    $\mathrm{56.95}$  \\ 
					
					\hline  
					\multirow{2}{*}{OURS }
					& $0.05d$ &
					 $\mathbf{82.20}$ &
					  $\mathbf{57.13 }$ &
					   $\mathbf{71.73}$ & 
					   $\mathbf{67.18}$
					    \\ \cline{2-6}
					\multirow{2}{*}{ }&
					$0.1d$ &
					$\mathbf{84.80}$ &
					 $\mathbf{60.02 }$ &
					  $\mathbf{73.23}$ &
					   $\mathbf{69.11}$\\ 
					
					% \hline
					\bottomrule
		\end{tabular}}}
		
	\end{table}

	\subsection{Recognition results on the public dataset}
	\label{sec:Effectofpublicdataset}
	To demonstrate not only the high recognition rate of our algorithm for complex and symmetric objects (e.g.spine) but also the effectiveness of our algorithm for objects of other shapes, we tested it under the public datasets UWA and DTU.

	Table \ref{tab:4} shows the recognition results of our algorithm and the other seven algorithms on the UWA dataset. In terms of time consumption, the time consuming of our algorithm is superior to the other algorithms except for the commercial software Halcon. In terms of recognition accuracy, we achieve a $ 100\% $ recognition rate for most objects, surpassing the other compared algorithms even in the highly occluded case. As shown in Fig. \ref{fig:uwadataset} for the qualitative comparison of the UWA dataset, it can be seen that our algorithm still has stable and correct results in the case of high occlusion.

	The DTU dataset contains many different types of geometric structure models. In order to more clearly show the effect of our algorithm on different geometric structures, we artificially divided the DTU dataset into geometrically complex, planar, and cylindrical by geometric structure.
	 The geometric classification of DTU dataset is available in Appendix.

	We selected some complex and symmetric objects with bone properties from the DTU dataset . The quantitative comparison results of these eight algorithms are shown in Table \ref{tab:DTU_SPINE}, which shows the clear advantage of our algorithm for this type of object.

We compare our algorithm with other algorithms for different geometric structures in the DTU dataset. The final results are shown in Table \ref{tab:6}, and it can be seen that our algorithm outperforms other matching algorithms for various types of 3D models in the DTU dataset. Thus, our algorithm owns scalability. A qualitative comparison of the DTU dataset is shown in Fig. \ref{fig:dtudataset1}.

	\section{Conclusion}
	\label{sec:Conclusion}
	For the structural characteristics of the human spine, we propose a pose estimation method based on the edge-enhanced point pair features, which mainly consists of an edge-based sampling method and an edge-matching-based pose verification method.  We perform extensive tests on the pig spine dataset and open datasets. The experiments demonstrate that our method is suitable for automatic surgical navigation systems due to its high accuracy, robustness, and short time consumption. Moreover, our algorithm is completely based on depth information for point cloud registration to serve as an excellent solution to light shading in surgical scenes.

	\section{Declarations}
	
	\label{Availability of data and materials}
	\subsection{Availability of data and materials}

	Open datasets in the manuscript are from public repositories (https://roboimagedata.compute.dtu.dk/ and http://vision.deis.unibo.it/keypoints3d/ds/UWA.7z),
	The open-source comparison algorithms are from public repositories (https://www.mvtec.com/products/halcon and https://gitlab.com/caro-sdu/covis).

	\label{Competing interests}
	\subsection{Competing interests}
	
	The authors declare that they have no known competing financial interests or personal relationships that could have appeared to influence the work reported in this paper.

	\label{Funding}
	\subsection{Funding}
This paper is supported in part by the National Key Research and Development Program of China (2018AAA0102200), National Nature Science Foundation of China (62132021, 62102435, 61902419, 62002375, 62002376), Natural Science Foundation of Hunan Province of China (2021JJ40696), Huxiang Youth Talent Support Program(2021RC3071) and NUDT Research Grants (ZK19-30, ZK22-52).
	
	\label{Authors' contributions}
	\subsection{Authors' contributions}
	\textbf{Chenyi Liu:} Methodology, Writing Draft, Visualization, Results Analysis; 
	\textbf{Fei Chen:} Methodology, Supervision; 
	\textbf{Lu Deng:} Supervision. 
	\textbf{Jia Wang:} Supervision;
	\textbf{Renjiao Yi:} Supervision, Results Analysis;
	\textbf{Lintao Zheng:} Supervision;  
	\textbf{Chenyang Zhu:} Supervision, Results Analysis; 
	\textbf{Kai Xu:} Methodology, Supervision.

	\label{Acknowledgements}
	\subsection{Acknowledgements}
		
		We thank Jiazhao Zhang and Yuqin Lan for helpful discussions. This paper is supported in part by the National Key Research and Development Program of China (2018AAA0102200), National Nature Science Foundation of China (62132021, 62102435, 61902419, 62002375, 62002376), Natural Science Foundation of Hunan Province of China (2021JJ40696), Huxiang Youth Talent Support Program(2021RC3071) and NUDT Research Grants (ZK19-30, ZK22-52).
	\newpage
		
\section*{Appendix}	
	\label{Appendix}
\begin{figure*}[h!]
	\centering
	\includegraphics[width=1.0\textwidth]{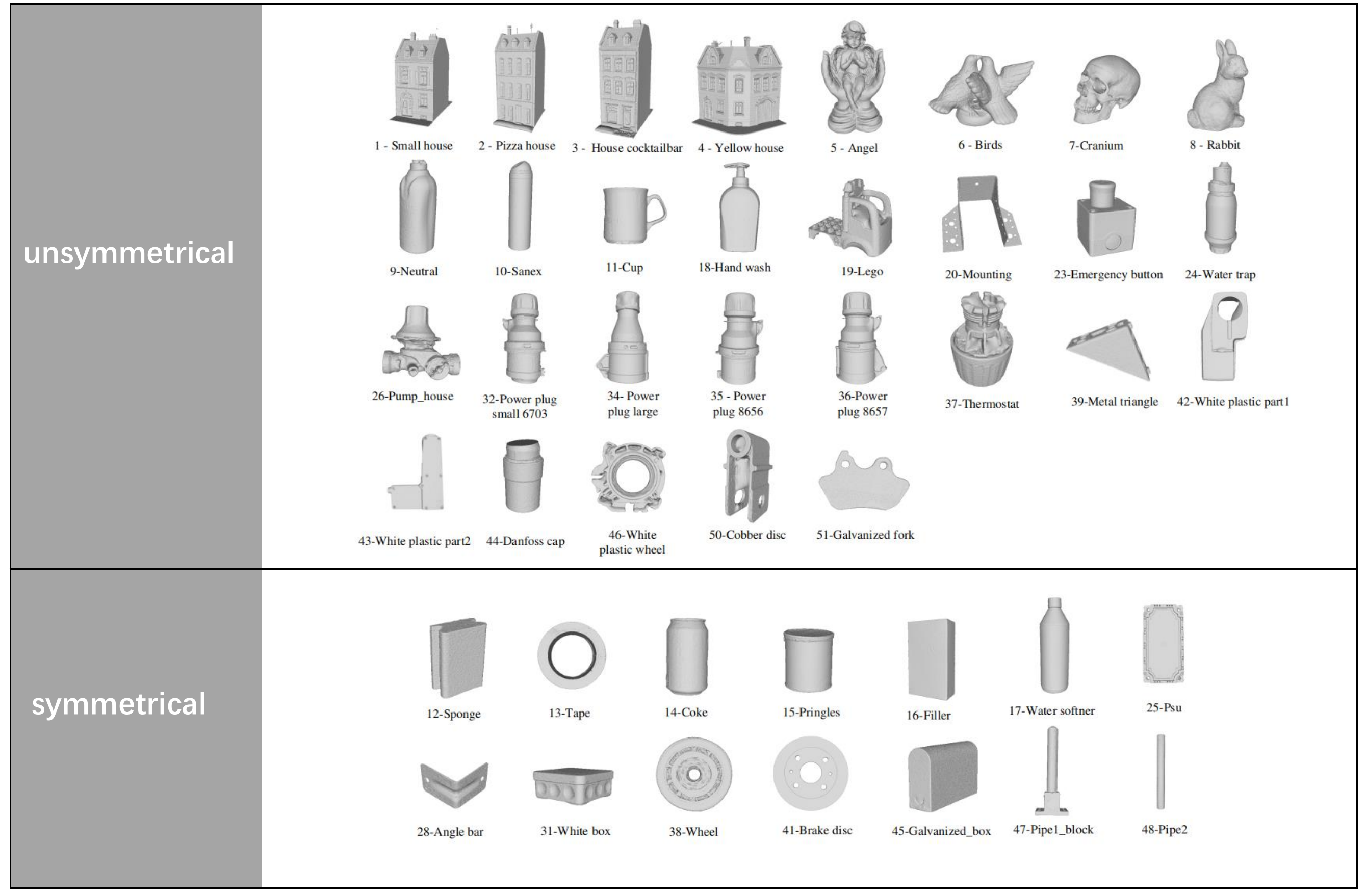}
	\caption{ \label{fig:symmetry} The symmetry classification of DTU dataset. }

\end{figure*}

\begin{figure*}[h!]
	\centering
	\includegraphics[width=1.0\textwidth]{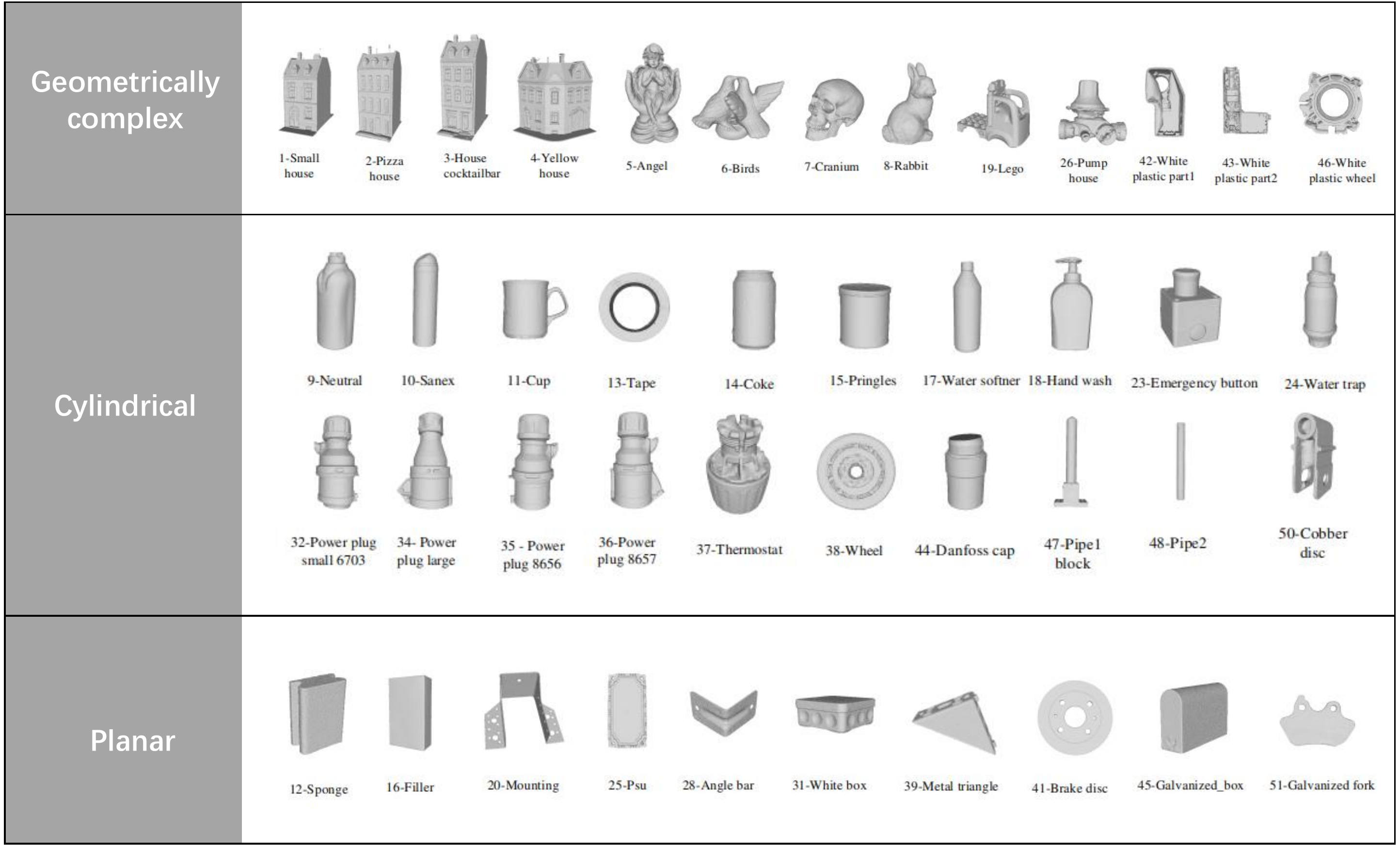}
	\caption{ \label{fig:shape} 	The geometric classification of DTU dataset. }
	
\end{figure*}

	As shown in  Fig \ref{fig:symmetry} and \ref{fig:shape},  We illustrate our classification of symmetric and geometric properties of DTU dataset \cite{solund2016large}.

	% for bibtex
	\bibliographystyle{CVMbib}
	\bibliography{refs}

\begin{thebibliography}{10}
\expandafter\ifx\csname urlstyle\endcsname\relax
  \providecommand{\doi}[1]{doi:\discretionary{}{}{}#1}\else
  \providecommand{\doi}{doi:\discretionary{}{}{}\begingroup
  \urlstyle{rm}\Url}\fi

\bibitem{li2019mixed}
Li R, Si W, Liao X, Wang Q, Klein R, Heng PA. Mixed reality based respiratory
  liver tumor puncture navigation. \emph{Computational Visual Media}, 2019,
  5(4): 363--374.

\bibitem{wang2021current}
Wang Y, Cao D, Chen SL, Li YM, Zheng YW, Ohkohchi N. Current trends in
  three-dimensional visualization and real-time navigation as well as
  robot-assisted technologies in hepatobiliary surgery. \emph{World journal of
  gastrointestinal surgery}, 2021, 13(9): 904.

\bibitem{kim2017vertebrae}
Kim K, Lee S. Vertebrae localization in CT using both local and global symmetry
  features. \emph{Computerized Medical Imaging and Graphics}, 2017, 58: 45--55.

\bibitem{5651280}
Rusu RB, Bradski G, Thibaux R, Hsu J. Fast 3D recognition and pose using the
  Viewpoint Feature Histogram. In \emph{2010 IEEE/RSJ International Conference
  on Intelligent Robots and Systems}, 2010, 2155--2162,
  \doi{10.1109/IROS.2010.5651280}.

\bibitem{marton2011combined}
Marton ZC, Pangercic D, Blodow N, Beetz M. Combined 2D--3D categorization and
  classification for multimodal perception systems. \emph{The International
  Journal of Robotics Research}, 2011, 30(11): 1378--1402.

\bibitem{6385874}
Madry M, Ek CH, Detry R, Hang K, Kragic D. Improving generalization for 3D
  object categorization with Global Structure Histograms. In \emph{2012
  IEEE/RSJ International Conference on Intelligent Robots and Systems}, 2012,
  1379--1386, \doi{10.1109/IROS.2012.6385874}.

\bibitem{johnson1999using}
Johnson AE, Hebert M. Using spin images for efficient object recognition in
  cluttered 3D scenes. \emph{IEEE Transactions on pattern analysis and machine
  intelligence}, 1999, 21(5): 433--449.

\bibitem{5152473}
Rusu RB, Blodow N, Beetz M. Fast Point Feature Histograms (FPFH) for 3D
  registration. In \emph{2009 IEEE International Conference on Robotics and
  Automation}, 2009, 3212--3217, \doi{10.1109/ROBOT.2009.5152473}.

\bibitem{tombari2010unique}
Tombari F, Salti S, Stefano LD. Unique signatures of histograms for local
  surface description. In \emph{European conference on computer vision}, 2010,
  356--369.

\bibitem{rusu2009detecting}
Rusu RB, Holzbach A, Beetz M, Bradski G. Detecting and segmenting objects for
  mobile manipulation. In \emph{2009 IEEE 12th International Conference on
  Computer Vision Workshops, ICCV Workshops}, 2009, 47--54,
  \doi{10.1109/ICCVW.2009.5457718}.

\bibitem{hinterstoisser2011multimodal}
Hinterstoisser S, Holzer S, Cagniart C, Ilic S, Konolige K, Navab N, Lepetit V.
  Multimodal templates for real-time detection of texture-less objects in
  heavily cluttered scenes. In \emph{2011 International Conference on Computer
  Vision}, 2011, 858--865, \doi{10.1109/ICCV.2011.6126326}.

\bibitem{besl1992method}
Besl PJ, McKay ND. {Method for registration of 3-D shapes}. In \emph{Sensor
  Fusion IV: Control Paradigms and Data Structures}, volume 1611, 1992, 586 --
  606, \doi{10.1117/12.57955}.

\bibitem{chen1992object}
Chen Y, Medioni G. Object modelling by registration of multiple range images.
  \emph{Image and vision computing}, 1992, 10(3): 145--155.

\bibitem{rusinkiewicz2001efficient}
Rusinkiewicz S, Levoy M. Efficient variants of the ICP algorithm. In
  \emph{Proceedings Third International Conference on 3-D Digital Imaging and
  Modeling}, 2001, 145--152, \doi{10.1109/IM.2001.924423}.

\bibitem{park2019pix2pose}
Park K, Patten T, Vincze M. Pix2pose: Pixel-wise coordinate regression of
  objects for 6d pose estimation. In \emph{Proceedings of the IEEE/CVF
  International Conference on Computer Vision}, 2019, 7668--7677.

\bibitem{hodan2020epos}
Hodan T, Barath D, Matas J. Epos: Estimating 6d pose of objects with
  symmetries. In \emph{Proceedings of the IEEE/CVF conference on computer
  vision and pattern recognition}, 2020, 11703--11712.

\bibitem{liang2019pointnetgpd}
Liang H, Ma X, Li S, Görner M, Tang S, Fang B, Sun F, Zhang J. PointNetGPD:
  Detecting Grasp Configurations from Point Sets. In \emph{2019 International
  Conference on Robotics and Automation (ICRA)}, 2019, 3629--3635,
  \doi{10.1109/ICRA.2019.8794435}.

\bibitem{lan2022arm3d}
Lan Y, Duan Y, Liu C, Zhu C, Xiong Y, Huang H, Xu K. ARM3D: Attention-based
  relation module for indoor 3D object detection. \emph{Computational Visual
  Media}, 2022: 1--20.

\bibitem{zeng2021ppr}
Zeng L, Lv WJ, Dong ZK, Liu YJ. PPR-Net++: Accurate 6-D Pose Estimation in
  Stacked Scenarios. \emph{IEEE Transactions on Automation Science and
  Engineering}, 2021.

\bibitem{drost2010model}
Drost B, Ulrich M, Navab N, Ilic S. Model globally, match locally: Efficient
  and robust 3D object recognition. In \emph{2010 IEEE Computer Society
  Conference on Computer Vision and Pattern Recognition}, 2010, 998--1005,
  \doi{10.1109/CVPR.2010.5540108}.

\bibitem{choi2016rgb}
Choi C, Christensen HI. RGB-D object pose estimation in unstructured
  environments. \emph{Robotics and Autonomous Systems}, 2016, 75: 595--613.

\bibitem{drost20123d}
Drost B, Ilic S. 3d object detection and localization using multimodal point
  pair features. In \emph{2012 Second International Conference on 3D Imaging,
  Modeling, Processing, Visualization \& Transmission}, 2012, 9--16.

\bibitem{liu2018point}
Liu D, Arai S, Miao J, Kinugawa J, Wang Z, Kosuge K. Point pair feature-based
  pose estimation with multiple edge appearance models (PPF-MEAM) for robotic
  bin picking. \emph{Sensors}, 2018, 18(8): 2719.

\bibitem{vock2019fast}
Vock R, Dieckmann A, Ochmann S, Klein R. Fast template matching and pose
  estimation in 3D point clouds. \emph{Computers \& Graphics}, 2019, 79:
  36--45.

\bibitem{lu39three}
Lu R, Zhu F, Wu Q, Chen F, Cui Y, Kong Y. Three-Dimensional Object Recognition
  Based on Enhanced Point Pair Features. \emph{Acta Optica Sinica}, 2019,
  39(8): 0815006.

\bibitem{vidal2018method}
Vidal J, Lin CY, Llad{\'o} X, Mart{\'\i} R. A method for 6D pose estimation of
  free-form rigid objects using point pair features on range data.
  \emph{Sensors}, 2018, 18(8): 2678.

\bibitem{guo2021efficient}
Guo J, Xing X, Quan W, Yan DM, Gu Q, Liu Y, Zhang X. Efficient Center Voting
  for Object Detection and 6D Pose Estimation in 3D Point Cloud. \emph{IEEE
  Transactions on Image Processing}, 2021, 30: 5072--5084.

\bibitem{hinterstoisser2016going}
Hinterstoisser S, Lepetit V, Rajkumar N, Konolige K. Going further with point
  pair features. In \emph{European conference on computer vision}, 2016,
  834--848.

\bibitem{papazov2010efficient}
Papazov C, Burschka D. An efficient ransac for 3d object recognition in noisy
  and occluded scenes. In \emph{Asian conference on computer vision}, 2010,
  135--148.

\bibitem{mian2006three}
Mian AS, Bennamoun M, Owens R. Three-dimensional model-based object recognition
  and segmentation in cluttered scenes. \emph{IEEE transactions on pattern
  analysis and machine intelligence}, 2006, 28(10): 1584--1601.

\bibitem{solund2016large}
S{\o}lund T, Buch AG, Kr{\"u}ger N, Aan{\ae}s H. A large-scale 3D object
  recognition dataset. In \emph{2016 Fourth International Conference on 3D
  Vision (3DV)}, 2016, 73--82.

\bibitem{hodavn2016evaluation}
Hoda{\v{n}} T, Matas J, Obdr{\v{z}}{\'a}lek {\v{S}}. On evaluation of 6D object
  pose estimation. In \emph{European Conference on Computer Vision}, 2016,
  606--619.

\bibitem{buch2017rotational}
Buch AG, Kiforenko L, Kraft D. Rotational subgroup voting and pose clustering
  for robust 3d object recognition. In \emph{2017 IEEE International Conference
  on Computer Vision (ICCV)}, 2017, 4137--4145.

\bibitem{jorgensen2015geometric}
J{\o}rgensen TB, Buch AG, Kraft D. Geometric Edge Description and
  Classification in Point Cloud Data with Application to 3D Object Recognition.
  In \emph{VISAPP (1)}, 2015, 333--340.

\bibitem{buch2016local}
Buch AG, Petersen HG, Kr{\"u}ger N. Local shape feature fusion for improved
  matching, pose estimation and 3D object recognition. \emph{SpringerPlus},
  2016, 5(1): 1--33.

\bibitem{salti2014shot}
Salti S, Tombari F, Di~Stefano L. SHOT: Unique signatures of histograms for
  surface and texture description. \emph{Computer Vision and Image
  Understanding}, 2014, 125: 251--264.

\end{thebibliography}
	
	\subsection*{Author biography}

	\begin{biography}[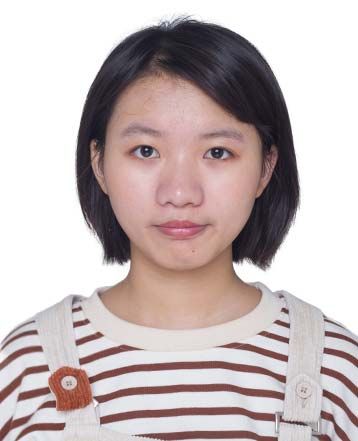]{Chenyi Liu} received her B.S. degree in Software engineering from Tianjin Normal University, China in 2020. She is now a master student at the National University of Defense Technology, China. Her research interests cover 3D point cloud registration.

	\end{biography}

	\begin{biography}	[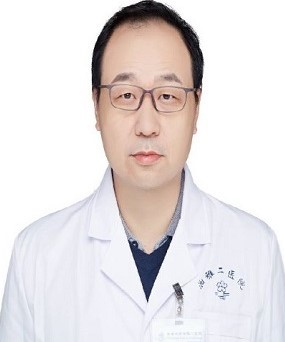]{Fei Chen} is a Professor at the spine surgery of the second xiangya hospital. The current direction of interest is  surgerical robot perception and  automatic navigation
	\end{biography}

\vspace*{3.6em}
\begin{biography}	[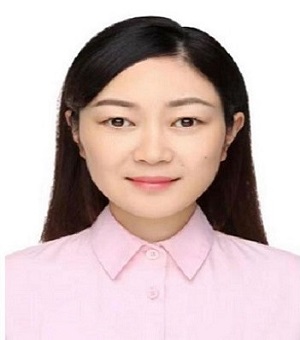]{Lu Deng}
	is a Professor at the surgery department of the second Xiangya hospital. The current direction of interest is automatic surgerical navigation.
\end{biography}

\vspace*{3.6em}
\begin{biography}	[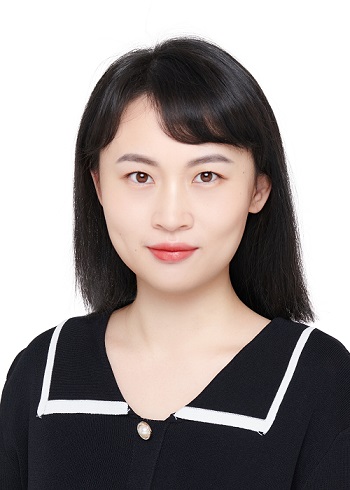]{Renjiao Yi}
	is an Assistant professor at the school of Computer, National University of Defense Technology.  She received her Ph.D. from Simon Fraser University in 2019. She is interested in 3D vision problems such as inverse rendering and image-based relighting. 
\end{biography}

\vspace*{3.6em}
\begin{biography}	[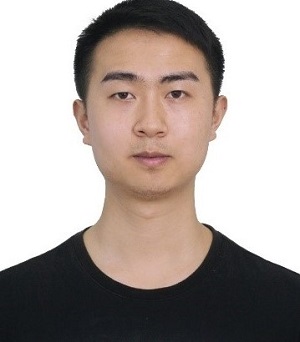]{Lintao Zheng}
	is an Assistant professor at College of Meteorology and Oceanography, National University of Defense Technology(NUDT). He earned his Ph.D in computer science from NUDT. His research interests focus on 3D vision and robot perception.
\end{biography}

\vspace*{3.6em}
\begin{biography}	[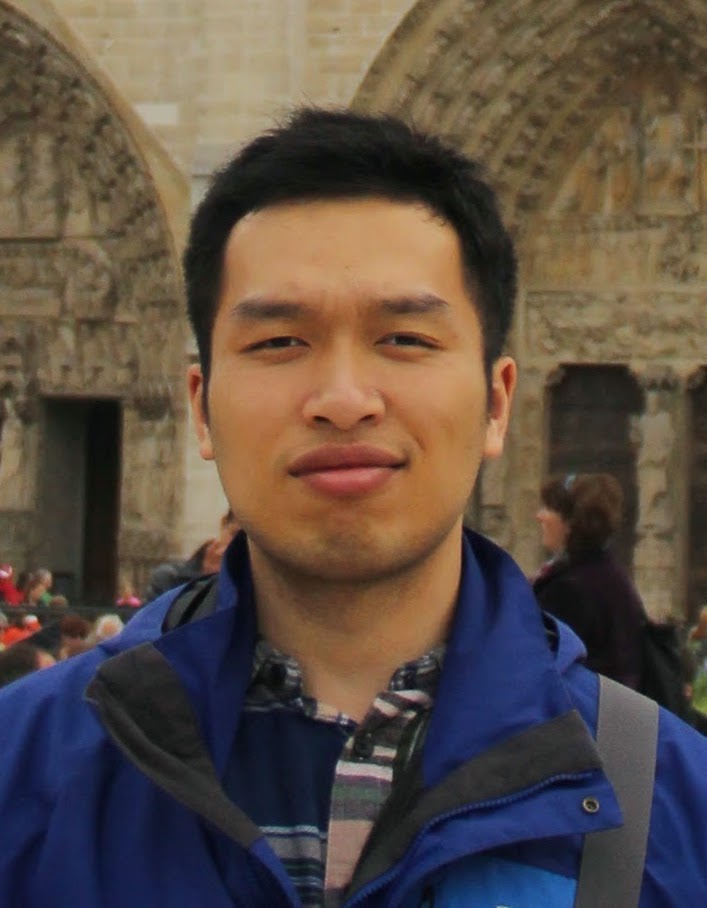]{Chenyang Zhu}
	is an Assistant Professor at School of Computer, National University of Defense Technology. He received his Ph.D. from Simon Fraser University in 2019. The current directions of interest include 3D vision and robot perception $\&$ navigation, etc.
\end{biography}

\vspace*{3.6em}
\begin{biography}	[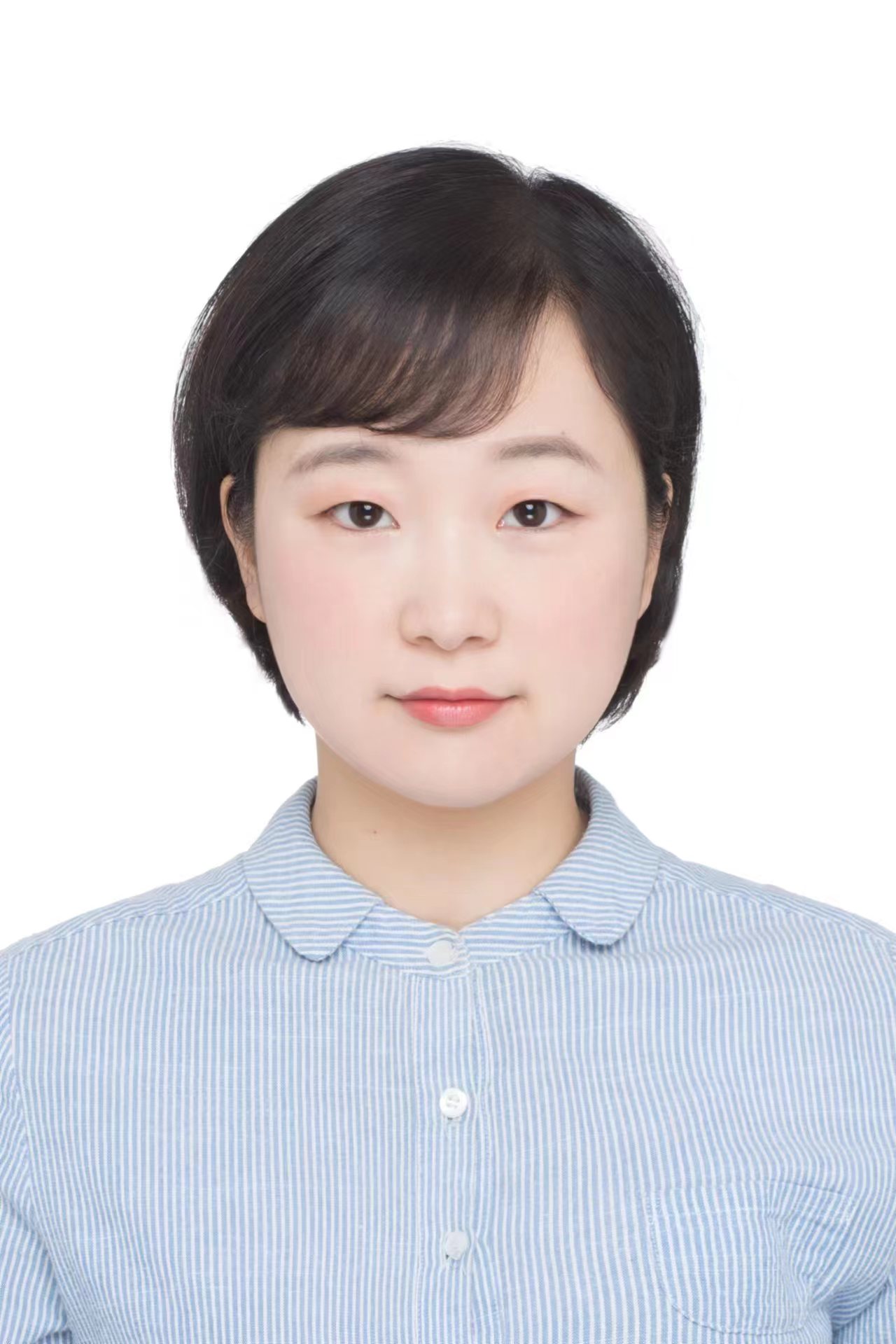]{Jia Wang} received her B.S. and M.E. degrees from National University of Defense Technology.  She is currently an assistant research fellow  with Beijing Institute of Tracking and Communication Technology. Her research interests focus on launch informatization.
\end{biography}

\vspace*{3.6em}
\begin{biography}	[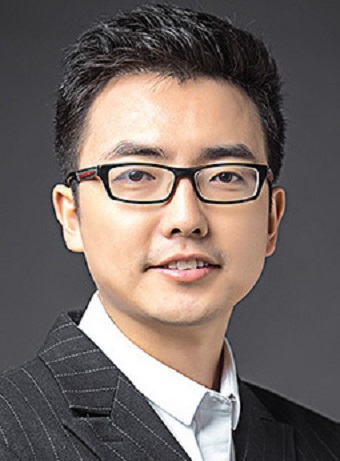]{Kai Xu} is a Professor at the School of Computer, National University of Defense Technology, where he received his Ph.D. in 2011. He serves on the editorial board of ACM Transactions on Graphics, Computer Graphics Forum, Computers $\&$ Graphics, etc.
\end{biography}

	\vspace*{2.6em}
	\subsection*{Graphical abstract}

		\begin{figure}[h!]
		\centering
		\includegraphics[width=0.5\textwidth]{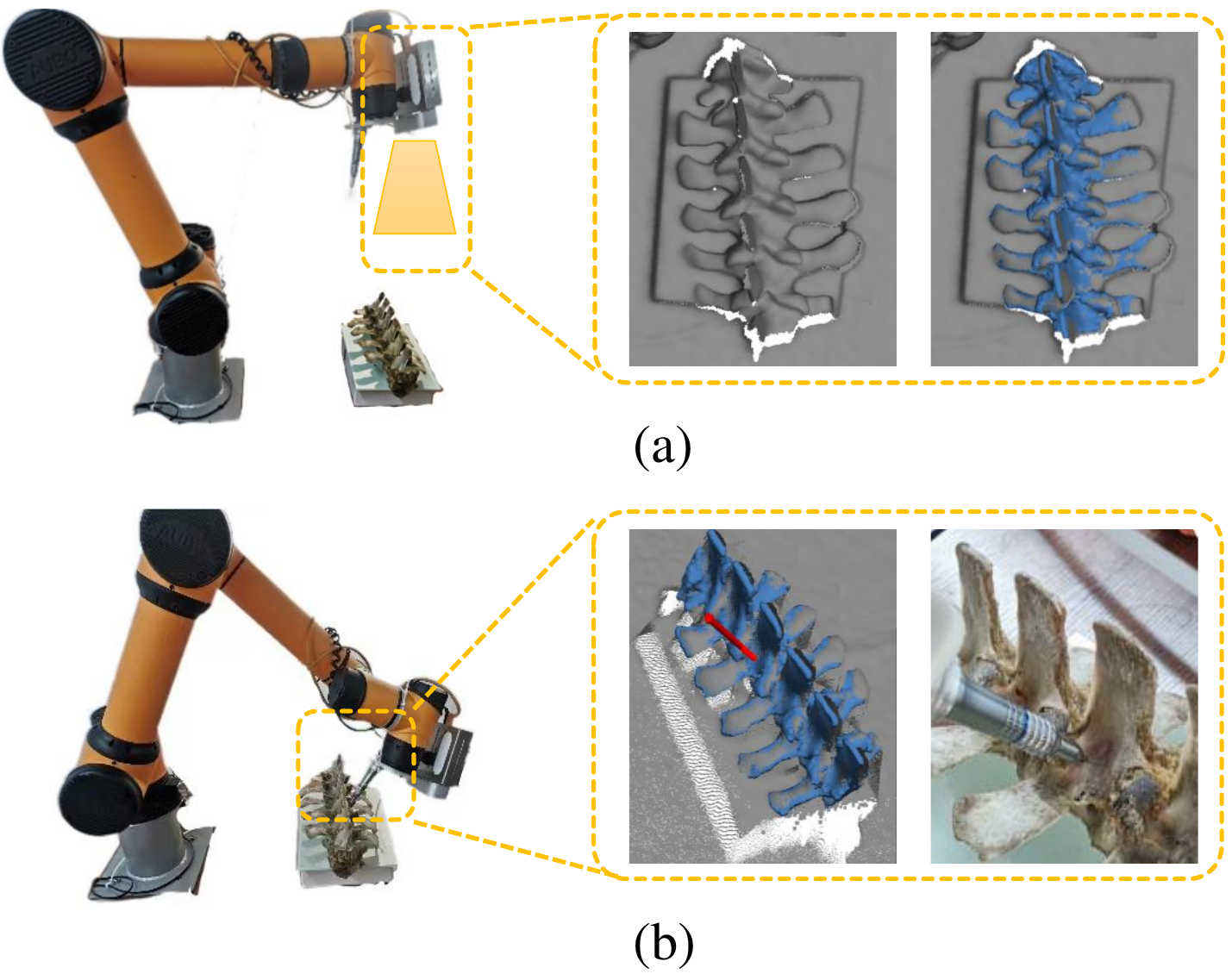}
		
	\end{figure}

\newpage

\end{document}